\documentclass[journal]{IEEEtran}
\usepackage{amsmath,amsfonts}
\usepackage{algorithmic}
\usepackage{algorithm}
\usepackage{array}
\usepackage[caption=false,font=normalsize,labelfont=sf,textfont=sf]{subfig}
\usepackage{textcomp}
\usepackage{stfloats}
\usepackage{url}
\usepackage{verbatim}
\usepackage{graphicx}
\usepackage{cite}
\usepackage{dsfont}
\usepackage[table,xcdraw]{xcolor}
\usepackage{hyperref} 
\usepackage{booktabs}
\usepackage{multirow}

\begin{document}

\title{\textit{SpikeAD}: Toward Multiplication-Free Spike-Controlled Neuromorphic Architecture for Energy-Efficient Small Object Detection}


\author{Zhifan Song, Haralampos-G. Stratigopoulos,~\IEEEmembership{Member,~IEEE}
\author{Zhifan Song and Haralampos-G. Stratigopoulos\\
\small{Sorbonne University, CNRS, LIP6, Paris, France}\vspace{-0.4cm}

 }
\thanks{Manuscript received April 19, 2021; revised August 16, 2021.}}



\maketitle

\begin{abstract}
Energy-constrained aerial platforms require accurate object detection under strict power and latency budgets, yet conventional inference remains bottlenecked by multiply-accumulate (MAC) operations. While spiking neural networks (SNNs) offer bio-inspired efficiency, many SNN detectors retain hybrid MACs in intermediate layers, or use spike-driven inference where multi-level spikes are interpreted as repeated binary events. We propose SpikeAD, a direct-trained spike-controlled neuromorphic aerial detector. In SpikeAD, sparse bounded integer spikes act strictly as multiplexer control signals for multiplier-free shift-add inference, preserving spike boundaries throughout a four-scale bidirectional feature pyramid. We introduce the Spike-Gated Multi-Scale (SGMS) block, performing receptive field expansion with activity-driven channel selection, implementable with adders and comparators. Evaluated on VisDrone and DOTA and across Nano, Small, and Medium variants, SpikeAD achieves up to 46.2\% mAP@50 on VisDrone (+4.8\% over SNN prior art), and generalizes to oriented bounding box (77.6\% mAP@50 on DOTA) using the proposed SpikeOBB head. Under the Horowitz 45nm arithmetic energy model following standard methodology in SNN literature, SpikeAD achieves up to 18.9$\times$ estimated reduction relative to an equivalent MAC-based implementation. We propose the Spike Processing Core (SPC), an RTL compute primitive to validate hardware realizability for reconfigurable and scalable custom accelerator chip design on FPGA/ASIC, highlighting a promising path toward next-generation ultra-low-power neuromorphic processors for aerial object detection. Link to the open-source and open-weight GitHub repository will be added upon publication.
\end{abstract}


\section{Introduction}
\IEEEPARstart{U}{nmanned} aerial vehicles (UAVs) are essential platforms for surveillance, search-and-rescue, environmental monitoring, and remote sensing~\cite{visdrone,dota}. These applications demand real-time onboard local inference to avoid the wireless communication costs, latency, and data feed privacy risks associated with cloud offloading. However, UAV platforms impose critical constraints on power consumption, weight, and thermal dissipation, making energy-efficient edge computing a necessary design requirement~\cite{biodrone,iscasdronesnn}.

Single-stage detectors such as the YOLO family~\cite{yolo,yolov10} remain the most practical choice for edge computing and UAV applications due to their speed-accuracy tradeoff~\cite{mcayolo,ednet}.  Two-stage detectors such as Faster-RCNN~\cite{fasterrcnn} are slower, and Transformer-based models~\cite{dino, rtdetr} achieve strong accuracy on standard benchmarks but are computationally prohibitive for edge deployment, and struggle with small and dense objects typical of aerial imagery. Recent efforts have optimized YOLO for aerial challenges~\cite{vamyolox,edgeyolo,mcayolo,ednet}. However, all these approaches rely on multiply-accumulate (MAC) operations. Even when deployed on NPUs, sustained MAC-based inference rapidly drains batteries and induces thermal throttling~\cite{ednet}, limiting UAV mission duration.

Spiking neural networks (SNNs) offer a brain-inspired, low-power alternative derived from biological neural systems~\cite{nature,ne}. By utilizing discrete spikes and leaky integrate-and-fire (LIF) dynamics~\cite{lif}, convolutions can be reduced to repeated weight additions, eliminating multiplier circuits. Combined with temporal sparsity, this enables order-of-magnitude energy savings on neuromorphic hardware~\cite{loihi,ncs}. 

Despite recent advances in directly trained SNN detectors~\cite{emsyolo,spikeyolo,metaspikeformer}, important gaps remain. First, state-of-the-art SNN detectors often rely on hybrid architectures where intermediate modules retain real-valued MAC operations~\cite{sfod,spikinghybridyolo,hybridspikingvit}; deploying them requires heterogeneous compute within the neural inference pipeline (interleaving neuromorphic chips and GPU/NPU)~\cite{towards}, with inter-chip data transfers that offset energy savings. Second, for models that utilize reparameterization, although these blocks can be algebraically fused at inference to enforce spike-driven execution, the fused operator removes internal spike boundaries and substantially expands the dense kernel footprint with higher parameter storage and operation counts. Finally, current models rely on a spike-driven assumption, where a multi-level integer spike necessitates repeated iterative accumulations~\cite{spikeyolo}. A unified aerial detection architecture that preserves bounded spike operands throughout the learned inference path while providing an explicit multiplier-free hardware mapping remains insufficiently explored.

To bridge these gaps, we propose SpikeAD (\textit{Addition/Aerial Detector}), an algorithmic, hardware-aware SNN architecture for aerial small target detection. SpikeAD introduces a spike-controlled inference path: 
each learned layer after the spike-encoding stem receives bounded integer spike tensors that act strictly as multiplexer control signals for shifted weight selection. Consistent with prior SNN studies, we evaluate arithmetic energy analytically, while the RTL core serves only to validate the realizability of the proposed operator rather than a complete accelerator. To circumvent the dense-kernel bloat of fused reparameterization, SpikeAD utilizes factorized spike mixers to preserve internal spike boundaries. A four-scale bidirectional feature pyramid is designed to enhance small-object feature fusion. We further introduce the Spike-Gated Multi-Scale (SGMS) block, which expands receptive fields with dilated parallel branches and performs dynamic channel selection via a binary spike gating mechanism requiring only adders and comparators. At the hardware abstraction level, we propose the Spike Processing Core (SPC) RTL primitive, providing a clear mapping from our algorithmic operators to inspire scalable custom neuromorphic chip design. 
The main contributions are:

\begin{itemize}
    \item \textbf{SpikeAD}: An energy-efficient neuromorphic aerial detector where the inference workload maps to multiplier-free, spike-controlled accumulation.
    \item \textbf{Spike-Gated Multi-Scale (SGMS) block}: A hardware-friendly multi-scale spatial mixer combining dilated depthwise convolutions with binary dynamic channel gating for enhanced, activity-driven feature extraction, implementable with adders and comparators.
    \item \textbf{Performance Validation}: SpikeAD supports both horizontal and oriented detection, achieving state-of-the-art SNN accuracy up to 46.2\% mAP@50 on VisDrone and 77.6\% mAP@50 on DOTA. Under a standardized arithmetic-energy model, it provides up to an 18.9$\times$ estimated reduction relative to the equivalent MAC workload.
    \item \textbf{Spike Processing Core (SPC)}: An RTL compute primitive exploiting multi-level spikes as control signals for shift-add routing, synthesized on Xilinx ZCU104 FPGA to establish a scalable foundation inspiring custom neuromorphic chip design. Full accelerator implementation and measurement remain future work.
\end{itemize}

\section{Background}
\paragraph{Aerial Object Detection and Edge Computing}
UAV-based object detection and remote sensing introduce unique challenges compared to natural image benchmarks~\cite{coco}, characterized by dense clustering, large-scale variations, and arbitrarily oriented targets (e.g. spanning 9.7-13.4\% of total image pixels~\cite{aodet}). Aerial and small object detection is an active research direction~\cite{tai_dota, tai_small}. While Transformer-based models (e.g., DINO~\cite{dino}, DETR~\cite{rtdetr}) achieve high accuracy, their computational overhead prohibits edge deployment.  Single-stage YOLO variants have been designed for aerial imagery~\cite{vamyolox,cslyolo,edgeyolo,ednet, mcayolo}. Optimized models such as EdgeYOLO~\cite{edgeyolo} and VAMYOLOX~\cite{vamyolox} target embedded GPUs, and EDNet~\cite{ednet} demonstrated mobile NPU inference. Furthermore, for remote sensing applications, oriented bounding box (OBB) detection is required. Specialized architectures such as PKINet~\cite{pkinet}, LSKNet~\cite{lsknet}, and Oriented R-CNN~\cite{orientedrcnn} employ multi-scale inception kernels and rotated proposals to capture directional targets. While FPGA-based CNN accelerators~\cite{fpgayolo1,fpgayolo2} have been proposed to reduce edge power consumption, all of the aforementioned architectures fundamentally rely on dense, floating-point or quantized MAC arrays to maintain accuracy. To the best of our knowledge, high-accuracy aerial and OBB detection has not yet been achieved using a fully multiplier-free neuromorphic paradigm.

\paragraph{Spiking Neural Networks and Hardware Acceleration}
Brain-inspired SNNs achieve energy efficiency through event-driven sparse additions~\cite{ncs, ne}. Deep SNN architectures have traditionally been constrained to image classification tasks~\cite{tianjic,ncsnnmnist}. Early adaptations for object detection, such as Spiking-YOLO~\cite{spikingyolo} and S3Det~\cite{s3det} (applied to DOTA), relied on ANN-to-SNN conversion. While functionally capable, conversion methodologies typically require hundreds to thousands of simulation timesteps to converge, rendering them impractical for real-time edge latency budgets. Direct training with surrogate gradients~\cite{surrogate,sfod} has enabled SNN detectors with low timesteps ($T{=}1$--$8$). Deep SNN architectures have progressed from vanilla spiking ResNets to advanced topologies that resolve gradient degradation through spike-element-wise shortcuts and membrane-based identity mappings~\cite{msresnet}. The spike-driven Transformer~\cite{metaspikeformer, espikeformer} reached high accuracy on ImageNet for classification tasks via spike-driven self-attention but incurs higher computational overhead with more parameters, higher energy, and lower accuracy compared to SpikeYOLO~\cite{espikeformer} in object detection tasks. 

Binary spikes suffer from quantization error in deep networks and integer-valued spikes~\cite{spikeyolo} mitigate this by encoding multi-bit information during training and converting back to spike-driven during inference, demonstrating accuracy comparable to YOLOv5. However, these advancements have exposed a critical hardware-algorithm deployment gap: the reparameterization dilemma. Unfused models retain real-valued intermediate operations, functioning as hybrid ANN-SNNs that necessitate heterogeneous hardware deployment~\cite{spikinghybridyolo,hybridspikingvit}. Alternatively, algebraic fusion enforces spike-driven execution but removes internal spike boundaries, shifting the computational burden from sparse pointwise mixing to dense convolution kernels, which offsets energy savings. Spiking-Hybrid-YOLO~\cite{spikinghybridyolo} targets ultra-low latency but achieves substantially lower accuracy; and~\cite{hybridspikingvit} demonstrated neuromorphic vision transformers for event camera detection. 

Furthermore, custom neuromorphic hardware implementations have been explored~\cite{snnfpga}, but mainly target image classification, and detection-capable implementations remain limited: they primarily target simple scenarios and small-scale datasets ~\cite{tcas1,iscasdronesnn,aicas}, while state-of-the-art SNN detectors~\cite{spikeyolo,metaspikeformer,sfod,spikingvit} report only theoretical energy estimates without providing a hardware-aware co-design path. SpikeAD addresses this gap by eliminating reparameterization reliance via spike-factorized mixers and introducing a spike-controlled compute abstraction, ensuring algorithmic advancements translate directly to highly efficient digital hardware execution.

\section{Methodology} 
\label{sec:methodology}

\begin{figure*}[htbp]
\centering 
\includegraphics[width=\textwidth]{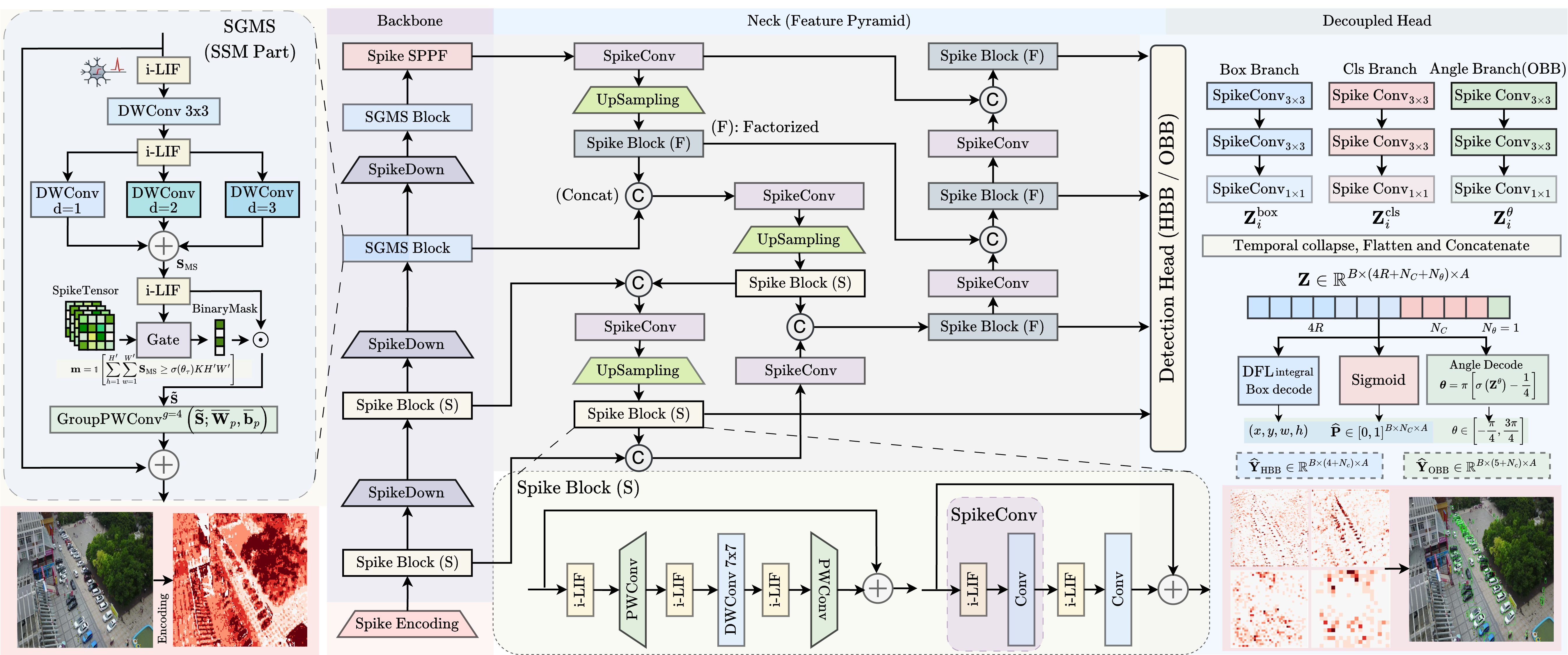} \caption{Architecture of SpikeAD. The spike encoding stem (CPU/Camera ISP) generates the initial spike tensor. The subsequent inference stage (accelerator-mapped), comprising the backbone, four-scale bidirectional feature pyramid neck, and decoupled head operates strictly via spike-controlled accumulation. The SpikeOBB block adds a dedicated orientation branch for oriented object detection. } 
\label{fig:spikead_arch} 
\vspace{-0.2cm}
\end{figure*} 

\subsection{Problem Formulation and Spike-Controlled Inference} 

As we operate on static-image datasets~\cite{visdrone, dota} rather than spike data from event cameras, the spike encoding stem performs conversion into the spike domain alongside $4\times$ strided downsampling, encoding each frame into bounded integer-valued spikes with maximum level $K=4$. We decouple the pipeline into pre-processing, inference, and post-processing following standard paradigms~\cite{ultralytics}. The encoding stem and the final bounding box decoding constitute lightweight pre- and post-processing steps. This is computationally negligible relative to the full model~\cite{addernet, dgrdet}, and is standard practice for SNNs on static-image data~\cite{emsyolo, spikeyolo}. In deployment, this stem maps naturally to the camera ISP or a host CPU, which is universally present in modern Systems-on-Chip (SoC)~\cite{heteroaisoc}. This allows the compute-heavy inference stage (backbone, neck, and head)~\cite{ultralytics} to execute exclusively on a dedicated spike-controlled accelerator.

Following recent SNN literature~\cite{espikeformer, spikeyolo}, SpikeAD adopts $T=1$ with multi-level spikes as an efficient operating point for static imagery. Its learned layers remain spike-mediated through surrogate-trained i-LIF neurons and discrete bounded outputs. SpikeAD uses each bounded spike as a control operand rather than interpreting its magnitude as repeated binary events. The 5D tensor representation is retained for neuromorphic compatibility for multi-timestep training comparison and suitability for future extension on direct fine-tuning on event-camera datasets. 

The architecture of SpikeAD is shown in Fig.~\ref{fig:spikead_arch}, and we discuss the constituent elements in the following sections.

\subsubsection{i-LIF Neuron} 

SpikeAD employs the integer-valued leaky integrate-and-fire (i-LIF) neuron to reduce activation quantization error and produce bounded integer spikes compatible with the proposed spike-controlled inference mechanism. Given an input $\mathbf{z}^{(t)}$, its membrane potential and spike output are defined as 
\begin{equation} 
\begin{aligned} 
\mathbf{u}^{(t)} &= 
\begin{cases} 
\mathbf{z}^{(t)}, & t=1,\\[1mm] \lambda \left( \mathbf{u}^{(t-1)}-\mathbf{s}^{(t-1)} \right) +\mathbf{z}^{(t)}, & t>1, 
\end{cases} 
\\ \mathbf{s}^{(t)} &= \mathcal{Q}_{K}\!\left(\mathbf{u}^{(t)}\right) = \text{round} \left[ \text{clip} \left( \mathbf{u}^{(t)},0,K \right) \right], 
\end{aligned} 
\label{eq:ilif} 
\end{equation} where $\lambda\in(0,1)$ is the membrane decay factor and $\mathbf{s}^{(t)}\in\{0,\ldots,K\}$ is the bounded integer spike output. During training, the derivative of $\mathcal{Q}_{K}$ is approximated by the straight-through estimator (STE) 
\begin{equation} 
\frac{\partial \mathcal{Q}_{K}(u)}{\partial u} \approx \mathds{1}\!\left[0\leq u\leq K\right]. \label{eq:ilif_ste} 
\end{equation} For compactness, $\mathcal{Q}_{K}(\cdot)$ subsequently denotes the complete i-LIF operation in Eq.~\eqref{eq:ilif}, including membrane integration, reset, and bounded integer quantization.

\subsubsection{Spike-Controlled Convolution} 

Let $\mathbf{S}^{(t)}=\mathcal{Q}_{K}(\mathbf{F}^{(t)})$ denote an integer spike feature map, with $\mathbf{S}^{(t)}\in\{0,\ldots,K\}^{B\times C_{\mathrm{in}}\times H\times W}$ and let $\mathbf{W}\in \mathbb{R}^{C_{\mathrm{out}}\times C_{\mathrm{in}}\times k\times k}$ denote a convolution kernel. For timestep $t$, the output at channel $j$ and spatial position $(h,w)$ is 

\begin{equation} 
y_{j}^{(t)}(h,w) = b_j + \sum_{c=1}^{C_{\mathrm{in}}} \sum_{(p,q)\in\Omega_k} \mathcal{R}_{K} \left( S_{c}^{(t)}(h+p,w+q), W_{jcpq} \right), \label{eq:spike_controlled_conv} \end{equation}
where $\Omega_k$ denotes the $k\times k$ kernel support. Since $S_{c}^{(t)}(h,w)\in\{0,\ldots,K\}$, multiplication by a spike can be expressed as a bounded weight-selection function 
\begin{equation} \mathcal{R}_{K}(s,w) = sw \in \{0,w,\ldots,Kw\}, \quad s\in\{0,\ldots,K\}. \label{eq:weight_selection} \end{equation}

Eq.~\ref{eq:weight_selection} defines the mathematical operation, not its hardware realization. Conventional spike-driven execution realizes $\mathcal{R}_{K}(s,w)$ through $s$ repeated accumulations or virtual binary substeps. In contrast, SpikeAD introduces a spike-controlled execution paradigm. Because the integer spike $s \in \{0,\ldots,4\}$  is bounded, it serves as a control signal to a multiplexer routing pre-computed weight multiples into the accumulator via wiring shifts in fixed-point datapath and exponent adjustment in floating-point realization. Consequently, every non-zero spike triggers exactly one accumulation regardless of its magnitude, decoupling algorithmic energy from the spike level. The hardware abstraction enabling this datapath, the RTL primitive (Spike Processing Core), is detailed in Section~\ref{sec:hardware}.

\subsubsection{Inference-Time Parameter Folding} 

Each training-time convolution is followed by batch normalization (BN). Let $\mathbf{W}$ and $\mathbf{b}$ denote the trained weights and bias, and let $\boldsymbol{\alpha}$, $\boldsymbol{\beta}$, $\boldsymbol{\mu}$, and $\boldsymbol{\sigma}^{2}$ denote the BN scale, offset, running mean, and running variance, respectively. The normalization is absorbed into the preceding convolution as \begin{equation} 
\widehat{\mathbf{W}} = \frac{\boldsymbol{\alpha}} {\sqrt{\boldsymbol{\sigma}^{2}+\epsilon}} \odot\mathbf{W}, \quad \widehat{\mathbf{b}} = \boldsymbol{\beta} + \frac{\boldsymbol{\alpha}} {\sqrt{\boldsymbol{\sigma}^{2}+\epsilon}} \odot \left( \mathbf{b}-\boldsymbol{\mu} \right).
\label{eq:bn_folding} 
\end{equation} where channel-wise factors are broadcast over the kernel dimensions. This one-time reparameterization eliminates normalization operations from the inference graph without changing its functional mapping. For bias-free training-time convolutions, $\mathbf{b}$ is initialized as the zero vector before folding.

\subsubsection{SpikeConv Primitive} 
We define SpikeConv (SC) as the composition of an i-LIF activation and a convolution with inference-folded parameters: 
\begin{equation} 
\textbf{SC} \left( \mathbf{F}; \widehat{\mathbf{W}}, \widehat{\mathbf{b}} \right) = \text{Conv} \left( \mathcal{Q}_{K}(\mathbf{F}); \widehat{\mathbf{W}}, \widehat{\mathbf{b}} \right), \label{eq:spikeconv} 
\end{equation} Since every convolution receives a bounded integer spike tensor, SC maps directly to the spike-controlled accumulation in Eq.~\eqref{eq:spike_controlled_conv}. We instantiate three basic forms: 
\begin{itemize} 
    \item $\operatorname{SC}_{1{\times}1}$: Spike pointwise convolution ; 
    \item  $\operatorname{SC}_{DW,k}$: Spike depthwise convolution with kernel size $k$; 
    \item $\operatorname{SC}_{\downarrow2}$: strided SpikeConv for spatial downsampling by two (SpikeDown).
\end{itemize}

\subsection{SpikeAD Architecture and Design Principles} 
SpikeAD comprises three principal parts: a hierarchical spiking backbone following a lightweight RGB stem, a four-scale bidirectional feature pyramid neck, and a decoupled spiking detection head. 

\begin{figure}[t]
\centerline{\includegraphics[width=\linewidth]{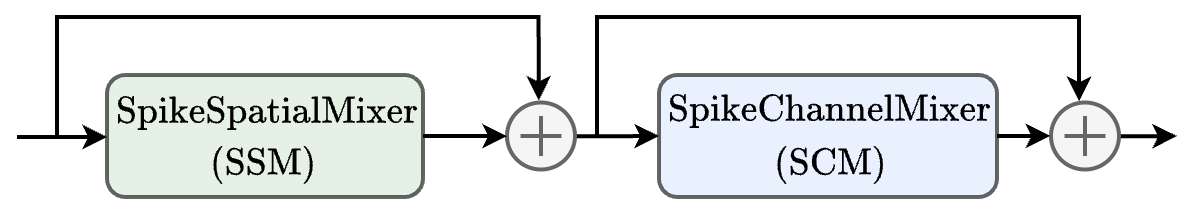}}
\caption{Structure of the Spike MetaFormer Block (shortened as SpikeBlock in the paper), comprising a Spike Spatial Mixer and a Spike Channel Mixer with successive residual connections.}
\label{fig:metaspikeblock}
\end{figure}

The backbone constructs four hierarchical feature levels 
\begin{equation} \mathbf{F}^{(\ell)} \in \mathbb{R}^{T\times B\times C_\ell\times \frac{H}{2^{\ell+1}}\times \frac{W}{2^{\ell+1}}}, \qquad \ell\in\{1,2,3,4\}. \label{eq:backbone_features} 
\end{equation} 
Each stage contains a sequence of Spike Blocks (MetaFormer style, Fig.~\ref{fig:metaspikeblock}), while adjacent stages are connected by strided SpikeConv downsampling. The four-scale bidirectional feature pyramid neck combines top-down semantic propagation and bottom-up feature aggregation. Unlike conventional three-scale detectors, it preserves the high-resolution level needed for tiny objects in aerial scenes, while deeper levels provide progressively larger receptive fields for medium and large objects.

SpikeAD follows four core design principles: spike continuity, factorized spatial--channel processing, multi-scale and high-resolution feature extraction and fusion, and direct mapping of operands to multiplier-free spike-controlled hardware primitives. First, every learned convolution after input encoding receives a bounded integer spike tensor. Second, the MetaFormer abstraction separates spatial and channel transformations, allowing each mixer to be optimized for spike-controlled execution with no hybrid interleaving or reparameterization that increases weight storage. Third, a spike-gated multi-scale feature extraction block and a four-scale bidirectional feature pyramid retains fine-grained spike features for dense aerial targets. Finally, all network operators are strictly constrained to map directly to the proposed multiplier-free SPC.

\subsection{Spike MetaFormer Block} 
\label{sec:spikeblocks}
Following the MetaFormer abstraction~\cite{poolformer}, the importance in vision tasks is to use Transformer-style blocks with a spatial and channel mixer with successive residual connection (Fig.~\ref{fig:metaspikeblock}, each Spike MetaFormer Block separates spatial token interaction from channel projection), as demonstrated~\cite{spikeyolo, metaspikeformer}, spiking self-attention achieves lower accuracy in object detection tasks, thus we employ convolution-based spatial mixer. Similar to token mixers in MetaFormer, in our case, we introduce spike mixers. Given an input spike feature map $ 
\mathbf{F} \in \mathbb{R}^{T\times B\times C\times H'\times W'}$,
the block is defined compactly as 
\begin{equation} 
\begin{aligned} \widetilde{\mathbf{F}} &= \mathbf{F} + \text{SSM} \left( \mathbf{F} \right), \\ \mathbf{F}' &= \widetilde{\mathbf{F}} + \text{SCM} \left( \widetilde{\mathbf{F}} \right), 
\end{aligned}
\label{eq:spike_meta_block} \end{equation} where $\text{SSM}$ and $\text{SCM}$ denote the Spike Spatial Mixer and Spike Channel Mixer, respectively. Both mixers are constructed exclusively from SpikeConv primitives, ensuring that every learned convolution receives integer-valued spikes.

\subsubsection{Spike Spatial Mixer} The SSM employs an inverted depthwise-separable topology~\cite{mobilenetv2} with expansion ratio $e=2$ and depthwise kernel size 7. It first expands the channel dimension, applies large-kernel depthwise spatial mixing, and projects the features back to the original dimension: 
\begin{equation}
    \text{SSM}(\mathbf{F}) = \underbrace{\textbf{SC}_{1{\times}1}^{eC\rightarrow C}}_{\text{compress}} \circ\ \underbrace{\textbf{SC}_{\text{DW},\, k}}_{\text{spatial}} \circ\ \underbrace{\textbf{SC}_{1{\times}1}^{C\rightarrow eC}}_{\text{expand}}(\mathbf{F})
    \label{eq:ssm}
\end{equation} 
The pointwise layers perform channel expansion and compression, while the DWConv provides spatial interaction at linear complexity with respect to the channel dimension. The structure is depicted on the bottom of Fig.~\ref{fig:spikead_arch}. The proposed SSM ensures spike continuity contrary to SpikeYOLO's SepConv~\cite{spikeyolo} which requires MAC in the last stage with unfused graph or fused graph that increases parameters and operations while removing spike boundaries.

\subsubsection{Spike Channel Mixer (Standard)} The SCM expands and subsequently restores the channel dimension with ratio $r=4$. In high-resolution stages, where the channel dimension remains moderate, we use two standard spike convolutions: \begin{equation} \text{SCM}_{\mathrm{std}}(\mathbf{F}) = \textbf{SC}_{3\times 3}^{rC\rightarrow C} \circ \textbf{SC}_{3\times 3}^{C\rightarrow rC} (\mathbf{F}). \label{eq:scm_standard} 
\end{equation} 

\subsubsection{Spike Channel Mixer (Factorized)}
\label{sec:sfc}
For deeper stages, we propose Spike Factorized Convolutions (SFC) to replace standard dense projections. Unlike prior art that rely on reparameterized RepConv blocks which either act as hybrid MACs when unfused or bloat dense kernel storage and destroy spike boundaries when fused~\cite{spikeyolo}, SFC operates entirely within the integer-spike domain. SFC is composed entirely of the SpikeConv primitive (Eq.~\eqref{eq:spikeconv}) in the following arrangement: a dimension-preserving pointwise SpikeConv, a dimension-preserving $3\times 3$ depthwise SpikeConv to introduce localized spatial context into the channel projection, and a final pointwise SpikeConv that performs the actual channel transition. SFC allows zero-valued spikes to dynamically gate computation at every internal sub-operator without dense kernel expansion that also ensures hardware deployment consistency on the SPC primitive.
\begin{equation} 
\text{SFC}^{C_{\mathrm{in}}\rightarrow C_{\mathrm{out}}} = \textbf{SC}_{1\times 1}^{C_{\mathrm{in}}\rightarrow C_{\mathrm{out}}} \circ \textbf{SC}_{DW,3}^{C_{\mathrm{in}}\rightarrow C_{\mathrm{in}}} \circ \textbf{SC}_{1\times 1}^{C_{\mathrm{in}}\rightarrow C_{\mathrm{in}}}, \label{eq:spike_factorized_conv} 
\end{equation} 
The factorized channel mixer is then 
\begin{equation} 
\text{SCM}_{\text{fac}}(\mathbf{F}) = \text{SFC}^{rC\rightarrow C} \circ \text{SFC}^{C\rightarrow rC} (\mathbf{F}). 
\label{eq:scm_factorized}
\end{equation}

\subsection{Spike-Gated Multi-Scale Spatial Mixer} 

At deeper stages, SpikeAD replaces the standard SSM with the proposed Spike-Gated Multi-Scale (SGMS) mixer to improve varying-size small object feature extraction through parallel dilated depthwise convolutions and introduces activity-driven binary channel gating before pointwise channel mixing. This design enhances multi-scale context while suppressing weak spike channels through a hardware-friendly accumulation-and-comparison path. 

Given spike input $\mathbf{S}_{0} = \mathcal{Q}_K(\mathbf{F}) \in \{0,\ldots,K\}^{T\times B\times C\times H'\times W'}$, SGMS first applies a depthwise $3\times 3$ SpikeConv and then three parallel dilated branches with dilation rates $\mathcal{D}=\{1,2,3\}$:
\begin{equation} \mathbf{S}_{\mathrm{MS}} = \mathcal{Q}_{K} \left[ \sum_{d\in\mathcal{D}} \text{DWConv}_{3\times3}^{(d)} \left( \mathcal{Q}_{K} \left[ \text{DWConv}_{3\times3} \left( \mathbf{S}_{0} \right) \right] \right) \right], \label{eq:sgms_multiscale} \end{equation} The effective receptive fields are therefore $5\times5$, $7\times7$, and $9\times9$. The branch outputs are summed and re-quantized before gating. 

The binary channel gate is computed by comparing spatial spike activity with a learned threshold:
\begin{equation} \mathbf{m} = \mathds{1} \left[ \sum_{h=1}^{H'} \sum_{w=1}^{W'} \mathbf{S}_{\mathrm{MS}}(:,:, :,h,w) \geq \sigma(\theta_{\tau}) K  H'  W' \right]. 
\label{eq:channel_gate}
\end{equation}
where $\sigma(\theta_{\tau})$ is a learnable parameter per block broadcast over channels, and the indicator uses a hard-forward / sigmoid-backward ($\sigma(\cdot)$) STE. $\mathds{1}[\cdot]$ denotes the binary indicator equivalent to a Heaviside operation. The resulting gate is broadcast spatially and used as pass-or-suppress control:
\begin{equation}
\widetilde{\mathbf{S}}=\operatorname{Broadcast}(\mathbf{m})\odot\mathbf{S}_{\mathrm{MS}}.
\label{eq:sgms_mask}
\end{equation}

In hardware, this binary mask can be implemented as conditional forwarding rather than multiplication. Eq.~\eqref{eq:sgms_mask} represents a conditional channel-enable operation. The selected spike tensor is projected through a grouped pointwise convolution and added to the residual input: 

\begin{equation} 
\begin{aligned} \overline{\mathbf{W}}_{p} &= \gamma_s \frac{\boldsymbol{\alpha}_{p}} {\sqrt{\boldsymbol{\sigma}_{p}^{2}+\epsilon}} \odot\mathbf{W}_{p}, \\ \overline{\mathbf{b}}_{p} &= \gamma_s \left[ \boldsymbol{\beta}_{p} + \frac{\boldsymbol{\alpha}_{p}} {\sqrt{\boldsymbol{\sigma}_{p}^{2}+\epsilon}} \odot \left( \mathbf{b}_{p}-\boldsymbol{\mu}_{p} \right) \right], \\ \mathbf{Y} &= \mathbf{F} + \text{GroupPWConv}^{(g)} \left( \widetilde{\mathbf{S}}; \overline{\mathbf{W}}_{p}, \overline{\mathbf{b}}_{p} \right).
\end{aligned} 
\label{eq:sgms_fused_output} 
\end{equation} 

During training, this projection includes batch normalization and a learnable residual scale $\gamma_s$, which are folded into the weights$(\overline{\mathbf{W}}_{p}, \overline{\mathbf{b}}_{p})$ for inference. The grouped pointwise projection uses $g=4$ groups, reducing the projection cost from $C^{2}$ to $C^{2}/g$ spike-controlled accumulations per spatial position and offering a favorable balance between cost and representational capacity.

Consequently, the inference-time SGMS mixer implements multi-scale spatial mixing and activity-driven binary channel selection, grouped pointwise mixing, and residual refinement without normalization or residual-scale operations in the inference graph. Its gating branch contains only global spike accumulation, a learnable threshold comparison, and binary pass-or-suppress control. SGMS therefore provides an adaptive channel-selection mechanism while preserving the multiplier-free and spike-controlled execution constraints of SpikeAD.

The SSM part of SGMS is depicted on the left side of Fig.~\ref{fig:spikead_arch} For the complete SGMS block, the SCM part reuses the factorized spike channel mixer.

\subsection{Unified Spiking Decoupled Detection Head}
\label{sec:head}
SpikeAD supports horizontal and oriented bounding box (HBB and OBB) detection through a unified spiking decoupled head (Fig.~\ref{fig:spikead_arch}, right side). For each pyramid level, independent box, class and optional angle branches are built from SpikeConv layers, so all learned transformations in the head also receive integer spike operands. The temporal collapsing after the final prediction SpikeConv is considered as the post-processing pipeline executed on a CPU. The HBB mode contains independent box regression and classification branches. We further propose the SpikeOBB block, which introduces a third spike-controlled branch for orientation estimation. All learned transformations in the three branches operate on integer spike features, and the temporal dimension is collapsed only after the final prediction SpikeConv.

Let $\mathcal{P}=\{\mathbf{F}_{i}\}_{i=1}^{L}$, $\mathbf{F}_{i}\in \mathbb{R}^{T\times B\times C_i\times H_i\times W_i}$, and $L=4$ denote the four-scale spiking feature pyramid. 
For each scale $i$ and branch $q\in\{\mathrm{box},\mathrm{cls},\theta\}$, define the three-stage branch function
\begin{equation}
  h_i^{q}(\mathbf{F}) = \textbf{SC}_{1\times1}^{C_q\rightarrow O_q} \circ
  \textbf{SC}_{3\times3}^{C_q\rightarrow C_q} \circ
  \textbf{SC}_{3\times3}^{C_i\rightarrow C_q}(\mathbf{F}),
\end{equation}
where $O_{\mathrm{box}}=4R$, $O_{\mathrm{cls}}=N_c$, and $O_\theta=1$. $R=16$ is the number of discrete bins used by distributional box regression. The box branch therefore predicts $4R$ channels, corresponding to $R$ logits for each of the four boundary distances $(l,t,r,b)$, while the classification branch predicts $N_c$ object class logits. 

In SpikeOBB mode, an additional single-channel branch predicts one orientation logit per spatial location. The predictions from the box, class, and orientation branches are flattened spatially and concatenated across the four feature scales (i.e., $\mathbf{Z}^{q} = \underset{i=1,\ldots,L}{\operatorname{Concat}} \left[ \operatorname{Flatten} \left( \mathbf{Z}_{i}^{q} \right) \right]$ for $q\in\{\mathrm{box},\mathrm{cls},\theta\}$). The orientation output $\mathbf{Z}^{\theta}$ is enabled only in SpikeOBB mode. The process is detailed in Fig.~\ref{fig:spikead_arch} on the right side. We collapse the temporal dimension via simple averaging over the per-timestep outputs. The final horizontal and oriented bounding boxes are resolved on the CPU using standard Distribution Focal Loss (DFL) decoding and sigmoid activations. Because the spike-controlled head handles the heavy feature projection, this CPU-side post-processing incurs negligible computational overhead relative to the network's inference workload.

\section{Experimental Results}
\label{sec:results}
\begin{table*}[t]
\centering
\caption{Comparison on VisDrone-DET-val (640$\times$640, 10 classes). GMACs=GFLOPs/2. Energy computed per Horowitz model~\cite{horowitz} (45\,nm, FP32): $E_\text{MAC} = 4.6$\,pJ, $E_\text{AC} = 0.9$\,pJ. $\eta$ = arithmetic energy reduction factor vs.\ equivalent-MACs ANN. T$\times$K: timesteps $\times$ spike levels.}
\label{tab:visdrone}
\renewcommand{\arraystretch}{0.8}
\setlength{\tabcolsep}{9.8pt}
\small
\begin{tabular}{l|c c c c c|c c}
\hline
\textbf{Method} & \textbf{Params (M)} & \textbf{GMACs/GACs} & \textbf{T$\times$K} & \textbf{Energy (mJ)} & $\boldsymbol{\eta}\uparrow$ & \textbf{mAP$_{50}$} & \textbf{mAP$_{50\text{-}95}$} \\
\hline\hline
\multicolumn{8}{l}{\textit{Two-stage detectors}} \\
\hline
Faster R-CNN~\cite{fasterrcnn}   & 41.4 & 104.0  & -- & 478.4  & 1$\times$    & 39.1 & 23.6 \\
Cascade R-CNN~\cite{cascadercnn} & 69.3 & 118.0  & -- & 542.8  & 1$\times$    & 38.5 & 23.6 \\
\hline
\multicolumn{8}{l}{\textit{Transformer-based detectors}} \\
\hline
Sparse DETR~\cite{sparsedetr}    & 41.0 & 60.5  & -- & 278.3  & 1$\times$ & 42.5 & 27.3 \\
Deformable DETR~\cite{deformabledetr}    & 40.0 & 86.5  & -- & 397.9  & 1$\times$ & 42.2 & 27.1 \\
DINO~\cite{dino}                 & 47.6 & 139.5  & -- & 641.7  & 1$\times$ & 46.2 & 29.4 \\
\hline
\multicolumn{8}{l}{\textit{One-stage detectors}} \\
\hline
YOLOv8-N~\cite{ultralytics}    & 3.0  & 4.1 & -- & 18.9   & 1$\times$ & 32.3 & 18.7 \\
YOLOv8-M                   & 25.9 & 39.6  & -- & 181.9  & 1$\times$ & 41.9 & 25.4 \\
YOLOv10-S                  & 8.1  & 12.5  & -- & 57.3   & 1$\times$ & 38.5 & 22.9 \\
YOLOv12-N~\cite{yolov12}    & 2.6  & 3.8   & -- & 17.2   & 1$\times$ & 33.1 & 19.1 \\
YOLOv12-M                   & 20.2 & 35.5  & -- & 163.4  & 1$\times$ & 43.3 & 26.4 \\
\hline
VAMYOLOX-S~\cite{vamyolox} & 9.7  & 30.2  & -- & 138.8  & 1$\times$ & 34.2 & 20.2 \\
VAMYOLOX-M                  & 27.1 & 75.7 & -- & 348.2  & 1$\times$ & 38.2 & 23.2 \\
LWUAVDet-S~\cite{lwuavdet}  & 5.2 & 9.6 & -- & 44.1 & 1$\times$ & 41.4 & 24.7 \\
MCA-YOLO~\cite{mcayolo}  & 6.3 & 6.2 & -- & 28.3 & 1$\times$ & 37.7 & 19.8 \\
EdgeYOLO-Tiny~\cite{edgeyolo}  & 5.8 & 13.6 & -- & 62.6 & 1$\times$ & 38.5 & 21.8 \\
EdgeYOLO~\cite{edgeyolo}  & 41.2 & 63.2 & -- & 290.7 & 1$\times$ & 44.8 & 26.4 \\
EDNet-S~\cite{ednet}        & 9.3  & 18.6  & -- & 85.3   & 1$\times$ & 42.5 & 25.6 \\
EDNet-M                     & 19.1 & 44.2  & -- & 203.3  & 1$\times$ & 47.1 & 28.7 \\
\hline
\multicolumn{8}{l}{\textit{Spiking neural network detectors}} \\
\hline
EMS-YOLO~\cite{emsyolo}   & 25.3  & 7.7\,+\,21.1  & 4$\times$1 & 54.3$^\dagger$   & 2.7$\times$  & 22.5 & 10.9 \\ 
SpikeYOLO-N~\cite{spikeyolo}         & 13.2  & 3.8\,+\,13.0  & 1$\times$4 & 29.1$^\dagger$   & 5.4$\times$  & 36.3 & 20.9 \\
SpikeYOLO-S                          & 23.2  & 6.7\,+\,21.3  & 1$\times$4 & 49.9$^\dagger$   & 5.6$\times$  & 38.3 & 22.3 \\
SpikeYOLO-M                          & 45.8  & 13.6\,+\,44.7 & 1$\times$4 & 102.5$^\dagger$  & 5.6$\times$  & 41.4 & 24.4 \\
SpikeYOLO-N\,(Fused)$^\circ$         & 34.5  & 26.9  & 1$\times$4 & 24.3$^\dagger$   & 11.2$\times$ & 36.3 & 20.9 \\
SpikeYOLO-M\,(Fused)$^\circ$         & 128.5 & 96.7  & 1$\times$4 & 86.6$^\dagger$   & 11.4$\times$ & 41.4 & 24.4 \\
\hline 
\rowcolor{gray!20} \textbf{SpikeAD-N ($T=4$) (Ours)} & 13.5 & 45.4 & 4$\times$4 & 40.9$^\star$ & 4.8$\times$ & 40.9 & 24.6 \\ 
\rowcolor{gray!20} \textbf{SpikeAD-N (Ours)} & 13.5 & 11.9 & 1$\times$4 & \textbf{11.4}$^\star$ & \textbf{17.3}$\times$ & \textbf{39.3} & \textbf{23.2} \\ 
\rowcolor{gray!20} \textbf{SpikeAD-S (Ours)} & 23.4 & 20.2 & 1$\times$4 & \textbf{19.3}$^\star$ & \textbf{17.6}$\times$ & \textbf{42.5} & \textbf{25.2} \\  
\rowcolor{gray!20} \textbf{SpikeAD-M (Ours)} & 46.2 & 36.5 & 1$\times$4 & \textbf{34.8}$^\star$ & \textbf{18.9}$\times$ & \textbf{46.2} & \textbf{27.9} \\ 
\hline    
\end{tabular}  
\par\noindent{\scriptsize $^\dagger$\,Spike-driven inference. 
\quad
$^\star$\,Spike-controlled inference.          \quad 
$^\circ$\, RepConv fused to dense Conv (our adaptation for fair fully spike-driven baseline).  
}                     
\end{table*} 

\subsection{Datasets and Environment Setup}
We evaluate SpikeAD on two widely adopted aerial detection benchmarks. VisDrone~\cite{visdrone} is used for horizontal bounding box (HBB) small object detection in UAV imagery (resolution 640 $\times$ 640), while DOTA~\cite{dota} is employed for oriented bounding box (OBB) detection in remote-sensing scenes. For DOTA we use an overlap of 200 of 1024 $\times$ 1024 patches for pre-processing, common to related work~\cite{pkinet,lsknet}. Since no large-scale event-based aerial detection benchmark is available, we follow prior SNN detection methodologies and evaluate on frame-based datasets using direct spike encoding~\cite{spikeyolo,espikeformer}. This experimental setting effectively tests whether a low-timestep multi-level SNN can deliver competitive detection accuracy on dense targets while exposing sufficient spike sparsity for energy-efficient inference. Rather than attempting to surpass the absolute accuracy of heavy, compute-intensive ANNs, our objective is to demonstrate near-SOTA detection capabilities within a novel, highly energy-constrained neuromorphic paradigm. 

SpikeAD variants are direct-trained using the stochastic gradient descent optimizer with a base learning rate of 0.01 and a momentum of 0.937. Standard data augmentation techniques, including mosaic and affine transformations, are applied during training to improve generalization. We also train YOLO models, SpikeYOLO, EDNet under the exact same setup for fair comparison. We train from scratch on VisDrone and then perform transfer learning on the DOTA dataset. Model training and evaluations are done with 4$\times$ NVIDIA A100 (80 GB) GPUs, an Intel Xeon Gold 6300 CPU, and 128 GB of RAM. The software stack utilizes Python 3.11, PyTorch 2.5.1, and CUDA 11.8. For hardware-aware validation, the RTL development of the proposed Spike Processing Core (SPC) was implemented on Xilinx Zynq UltraScale+ MPSoC ZCU104 FPGA. Logic synthesis and place-and-route are performed using Vivado 2025.1.

\subsection{Energy Estimation Methodology}

Following prior SNN detection literature~\cite{metaspikeformer,espikeformer,spikeyolo,emsyolo}, we quantify the theoretical inference arithmetic energy using the Horowitz 45 nm CMOS model~\cite{horowitz}. To ensure a fair algorithmic baseline against conventional ANNs and prior SNNs, we adopt the floating-point 32-bit (FP32) precision, where a multiply-accumulate (MAC) operation consumes $E_{\text{MAC}} = 4.6$\,pJ and an accumulation consumes $E_{\text{AC}} = 0.9$\,pJ. This model isolates arithmetic computation and excludes memory accesses, data movement, routing, and control overheads. Therefore, the reported values represent estimated arithmetic energy rather than measured FPGA power or end-to-end system energy. For a conventional ANN executing $N$ equivalent operations, the energy is $E_{\text{ANN}} = N \cdot E_{\text{MAC}}$. In a conventional spike-driven SNN, an integer spike contributes proportionally to its magnitude, requiring repeated accumulations. Its energy is therefore $E_{\text{driven}} = \bar{s} \cdot N \cdot E_{\text{AC}} \propto \text{firing rate} $, where $\bar{s}$ is the operation-weighted mean spike value across the network. In contrast, our proposed Spike Processing Core (SPC) uses the spike strictly as a multiplexer control signal. Consequently, each nonzero spike triggers exactly one selected accumulation regardless of its magnitude. The energy for spike-controlled inference is therefore strictly bounded by the active spike rate $\rho$:
$E_{\text{controlled}} = \rho \cdot N \cdot E_{\text{AC}}$
where $\rho = P(s>0)$. The $s=3$ case requires one lightweight combinatorial pre-add, and we rigorously account for this by charging it at the full $E_{AC}$ cost, weighted by its precise activation frequency across the network. The total estimated FP32 energy is reported in Section~\ref{sec:compsota} to directly compare with prior work, while Section~\ref{sec:quant} details model quantization for true hardware deployment.

\subsection{Performance Comparison with State-of-The-Art}
\label{sec:compsota}

\begin{figure*}[htbp]
\centerline{\includegraphics[width=0.95\textwidth]{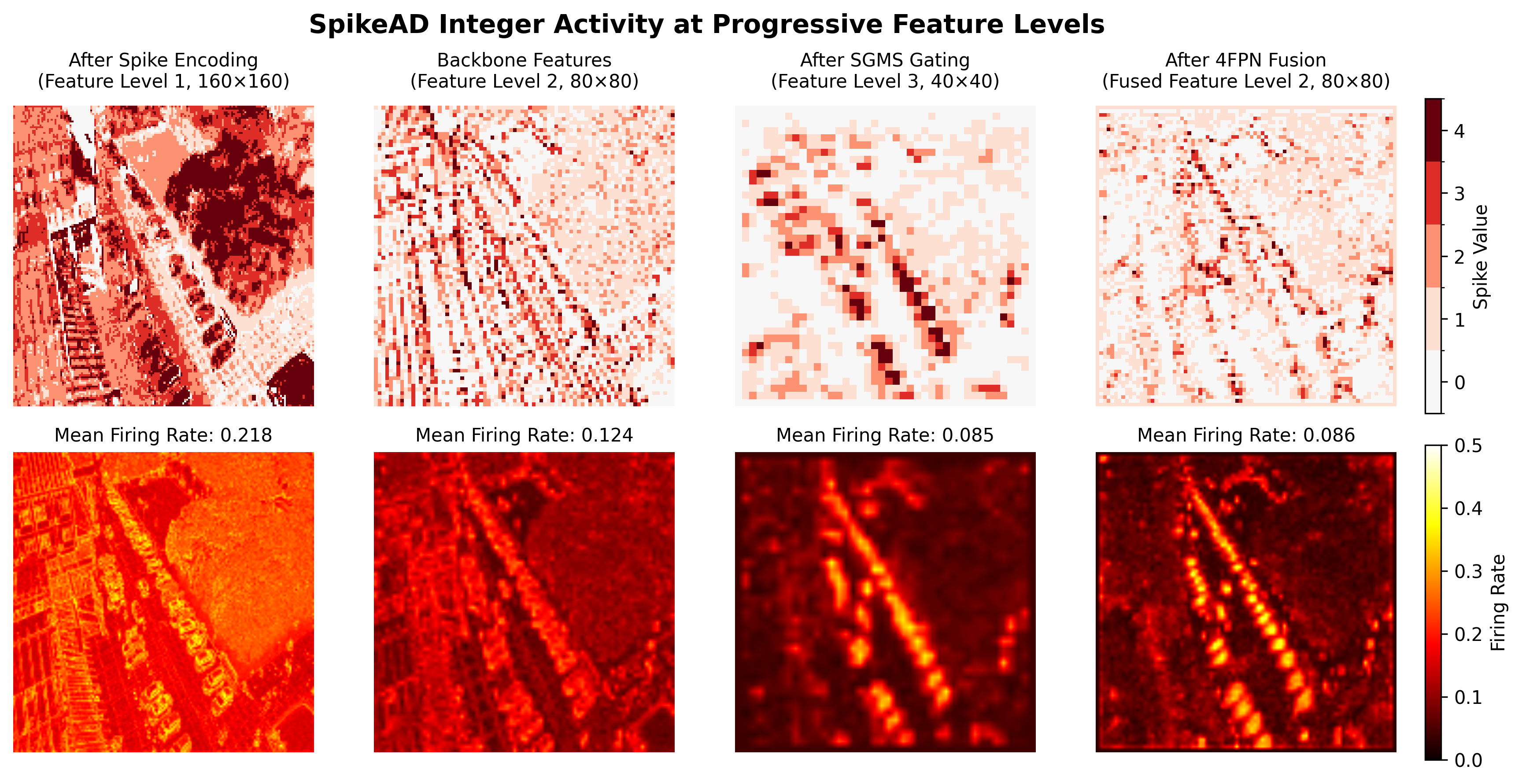}}\vspace{-0.6cm}
\caption{Integer spike feature maps and spatial firing rates at progressive network stages. Top: representative channel showing discrete spike values $\in \{0,1,2,3,4\}$. The channel with the most spike activity is selected. Bottom: spatial firing rate, averaged across time and channels. From left to right, we observe progressive abstraction: (a) fine-grained spatial encoding after spike encoding, (b) backbone feature development, (c) highly selective response after the SGMS gating mechanism, and (d) object-focused attention after FPN fusion.}
\label{fig:spike_hm}
\end{figure*}
\begin{figure}[t]
\centerline{\includegraphics[width=\linewidth]{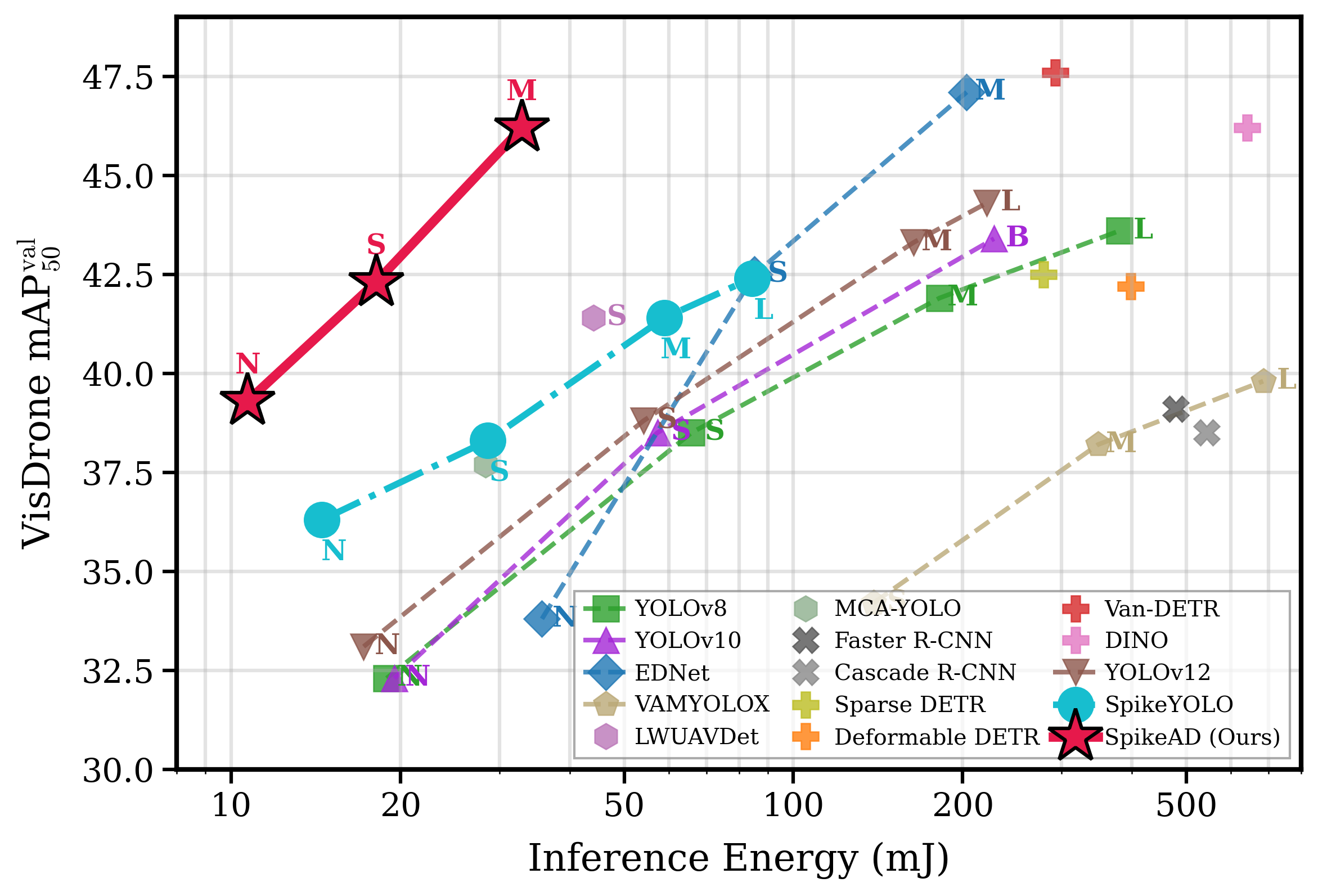}}\vspace{-0.2cm}
\caption{mAP vs. estimated arithmetic energy (Horowitz 45 nm model) curves: comparison with state-of-the-art methods on the VisDrone benchmark. }
\label{fig:fom}
\end{figure}

\subsubsection{Small Object Detection in Drone Imagery (VisDrone)}

We evaluate SpikeAD and conduct a comprehensive comparative evaluation on the VisDrone benchmark to compare with state-of-the-art ANN and SNN detectors. As shown in Table~\ref{tab:visdrone}, SpikeAD demonstrates a superior accuracy-energy tradeoff. Conventional two-stage architectures (e.g. Faster R-CNN~\cite{fasterrcnn}) and Transformer-based models (e.g. DINO~\cite{dino}) typically incur severe computational overhead, requiring hundreds of millijoules of estimated arithmetic energy, making them less suitable for resource-constrained UAV platforms. Compared to real-time one-stage ANNs, SpikeAD establishes highly competitive detection mAP operating with substantially lower estimated arithmetic energy: the M variant achieves 46.2\% mAP@50, outperforming conventional models such as YOLOv8-M (41.9\%), YOLOv12-M (43.3\%) and has an estimated 34.8 mJ of energy per inference, compared to 181.9 mJ and 163.4 mJ, respectively. 

Furthermore, we compare SpikeAD against recent ANNs tailored for UAV imagery, such as EDNet~\cite{ednet}, LWUAVDet~\cite{lwuavdet}, and MCA-YOLO~\cite{mcayolo}, which optimize spatial features for dense small targets, and models targeting edge computing such as EdgeYOLO~\cite{edgeyolo}. While EDNet-M achieves higher mAP@50 (47.1\%), its reliance on dense MAC arrays consume 203.3 mJ per frame. In contrast, SpikeAD-S achieves comparable detection performance (42.5\%) to EDNet-S (42.5\%) but operates at a fraction of the power (19.3 mJ vs. 85.3 mJ, 77.4\% reduction). This suggests that the proposed spike-controlled inference can match specialized aerial detectors without relying on dense MAC-based inference.

We also compare with the prior art SNN detectors. EMS-YOLO uses multi-timestep but binary spikes and the accuracy is relatively low for small object detection. For SpikeYOLO, we report two deployment interpretations. The first follows the unfused graph in released code, which retains real-valued intermediate operations and functions as a hybrid ANN-SNN. To establish a fully spike-driven baseline, we algebraically fused the reparameterized blocks (denoted as SpikeYOLO Fused$^\circ$). As discussed in Section~\ref{sec:spikeblocks}, while this enforces spike-driven execution, the dense kernel fusion removes internal spike boundaries and increases the parameter count from 45.8M to 128.5M (2.8$\times$) with no spike boundaries either. SpikeAD circumvents this dilemma entirely through its factorized mixers, outperforming both SpikeYOLO variants while maintaining low energy and parameter counts.

Among the evaluated models under the adopted arithmetic-energy methodology, SpikeAD provides a favorable accuracy-energy trade-off (Fig.~\ref{fig:fom}). SpikeAD-M surpasses SpikeYOLO-M by +4.8\% mAP@50 while simultaneously reducing energy consumption by approximately 3$\times$. To accommodate the diverse power and payload of constraints of varying UAV platforms, we introduced the N, S, M variants. Paired with the proposed SPC primitive, these variants offer highly scalable efficiency for custom accelerator designs based on specific needs for different drones. SpikeAD-N requires only 11.4 mJ per frame, a 17.3$\times$ arithmetic energy reduction against an equivalent MAC ANN, while maintaining a competitive 39.3\% mAP@50. As illustrated in the figure of merit (Fig~\ref{fig:fom}), which plots mAP@50 against arithmetic inference energy in log scale, SpikeAD variants lie on the Pareto frontier of the reported accuracy and arithmetic-energy results, indicating an improved trade-off relative to the evaluated SNN baselines.

While multi-step binary inference ($T{=}4$, $K{=}1$) degrades detection accuracy (down to 33.2\% mAP@50, not reported in Table), increasing integer-valued timesteps to $T{=}4$ on SpikeAD-N ($4{\times}4$) scales energy $3.8{\times}$ to 40.9\,mJ while achieving limited mAP gain over its $T{=}1$ counterpart. Moreover, SpikeAD-M achieves 46.2\% mAP$_{50}$ at 34.8\,mJ. Therefore, we adopt $T{\times K}{=}1{\times}4$ as our configuration. The results are consistent with prior observations on frame-based SNN detection on the COCO dataset~\cite{spikeyolo}. 

To provide interpretability and validate the energy efficiency of our spike-controlled architecture, Fig.~\ref{fig:spike_hm} visualizes the integer spike activity in feature maps and corresponding firing rates at progressive network stages. The top row shows representative channel activations restricted to discrete levels from 0 to 4, visually confirming that intermediate activations remain restricted to the discrete levels. At the initial stage after spike encoding, fine-grained scene details are preserved, yielding a relatively high mean firing rate (FR) of 0.218. As the features propagate through the backbone, a transition occurs after the proposed SGMS block: contrasting the heatmaps,  the spatial activity shifts drastically from dense textual extraction to a sparse, object-selective response, driving the local FR down to 0.085. This visually suggests the efficacy of the SGMS multi-scale feature extraction and binary gating mechanism in suppressing background noise. Following the high-resolution four-scale bidirectional FPN (HRFPN) fusion, the network demonstrates further refined, object-focused attention, clearly illuminating target structures of the cars (dominant objects in the specific frame, which is the same input shown in Fig.~\ref{fig:spikead_arch} and Fig.~\ref{fig:ablation}). Across the entire network, SpikeAD-N maintains an exceptionally low average FR of 10.5\% and an active spike rate $\rho$ of 27.8\%, while the M variant further suppresses the global FR to 9.8\%. 
\begin{table*}[t]
\centering
\caption{Comparison on DOTA-v1.0 test set. All models use single-scale training and testing. Per-class AP and mAP$_{50}$ are reported. $^\star$Spike-controlled inference (SPC). The abbreviations refer to Plane (PL), Baseball Diamond (BD), Bridge (BR), Ground Track Field (GTF), Small Vehicle (SV), Large Vehicle (LV), Ship (SH), Tennis Court (TC), Basketball Court (BC), Storage Tank (ST), Soccer-ball Field (SBF), Roundabout (RA), Harbor (HA), Swimming Pool (SP), and Helicopter (HC).}
\label{tab:dota}
\renewcommand{\arraystretch}{0.8}
\setlength{\tabcolsep}{2.0pt}
\scriptsize
\begin{tabular}{l|c|cc|ccccccccccccccc|c}
\hline
\textbf{Method} & \textbf{Params} & \textbf{GMACs/GACs} & \textbf{Energy(mJ)} & \textbf{PL} & \textbf{BD} & \textbf{BR} & \textbf{GTF} & \textbf{SV} & \textbf{LV} & \textbf{SH} & \textbf{TC} & \textbf{BC} & \textbf{ST} & \textbf{SBF} & \textbf{RA} & \textbf{HA} & \textbf{SP} & \textbf{HC} & \textbf{mAP} \\
\hline\hline
\multicolumn{20}{l}{\textit{Transformer-based oriented detectors}} \\
\hline
AO2-DETR~\cite{ao2detr}    & 40.8M & 152.0 & 699.2 & 87.99 & 79.46 & 45.74 & 66.64 & 78.90 & 73.90 & 73.30 & 90.40 & 80.55 & 85.89 & 55.19 & 63.62 & 51.83 & 70.15 & 60.04 & 70.91 \\
ARS-DETR~\cite{arsdetr}   & 41.6M & --  & -- & 86.61 & 77.26 & 48.84 & 66.76 & 78.38 & 78.96 & 87.40 & 90.61 & 82.76 & 82.19 & 54.02 & 62.61 & 72.64 & 72.80 & 64.96 & 73.79 \\
\hline
\multicolumn{20}{l}{\textit{One-stage oriented detectors}} \\
\hline
Rotated FCOS~\cite{fcos} & 31.9M & 103.5 & 476.1 & 88.52 & 77.54 & 47.06 & 63.78 & 80.42 & 80.50 & 87.34 & 90.39 & 77.83 & 84.13 & 55.45 & 65.84 & 66.02 & 72.77 & 49.17 & 72.45 \\
S2ANet~\cite{s2anet}          & 38.5M & 99.0 & 455.4 & 89.11 & 82.84 & 48.37 & 71.11 & 78.11 & 78.39 & 87.25 & 90.83 & 84.90 & 85.64 & 60.36 & 62.60 & 65.26 & 69.13 & 57.94 & 74.12 \\
O-RepPoints~\cite{oreppoints} & 36.6M & 97.0 & 446.2 & 87.02 & 83.17 & 54.13 & 71.16 & 80.18 & 78.40 & 87.28 & 90.90 & 85.97 & 86.25 & 59.90 & 70.49 & 73.53 & 72.27 & 58.97 & 75.97 \\
R3Det-KLD~\cite{r3detkld}     & 41.9M & 168.0 & 772.8 & 88.90 & 84.17 & 55.80 & 69.35 & 78.72 & 84.08 & 87.00 & 89.75 & 84.32 & 85.73 & 64.74 & 61.80 & 76.62 & 78.49 & 70.89 & 77.36 \\
\hline
\multicolumn{20}{l}{\textit{Two-stage oriented detectors}} \\
\hline
RoI Trans~\cite{roitrans}     & 55.1M  & 100.0    & 460.0 & 89.01 & 77.48 & 51.64 & 72.07 & 74.43 & 77.55 & 87.76 & 90.81 & 79.71 & 85.27 & 58.36 & 64.11 & 76.50 & 71.99 & 54.06 & 74.05 \\
MAK-BRNet-R~\cite{makbrnet}                   & 46.0M & 65.5 & 301.3 & 89.22 & 83.52 & 51.25 & 67.40 & 76.78 & 81.07 & 87.13 & 90.87 & 87.48 & 86.23 & 59.07 & 64.14 & 72.48 & 69.92 & 71.09 & 75.84 \\
Oriented R-CNN~\cite{orientedrcnn}           & 41.1M  & 99.5    & 457.7 & 89.46 & 82.12 & 54.78 & 70.86 & 78.93 & 83.00 & 88.20 & 90.90 & 87.50 & 84.68 & 63.97 & 67.69 & 74.94 & 68.84 & 52.28 & 75.87 \\
LSKNet-S~\cite{lsknet}        & 31.0M  & 80.5    & 370.3 & 89.66 & 85.52 & 57.72 & 75.70 & 74.95 & 78.69 & 88.24 & 90.88 & 86.79 & 86.38 & 66.92 & 63.77 & 77.77 & 74.47 & 64.82 & 77.49 \\
PKINet-S~\cite{pkinet}     & 30.8M  & 92.0    & 423.2 & 89.72 & 84.20 & 55.81 & 77.63 & 80.25 & 84.45 & 88.12 & 90.88 & 87.57 & 86.07 & 66.86 & 70.23 & 77.47 & 73.62 & 62.94 & 78.39 \\
\hline
\multicolumn{20}{l}{\textit{SNN-based oriented detectors}} \\
\hline
S3Det-50~\cite{s3det} & 121.1M & 104.0 & 93.6 & 87.83 & 77.45 & 34.05 & 64.50 & 62.34 & 73.10 & 66.11 & 90.56 & 75.33 & 78.28 & 55.37 & 56.12 & 62.88 & 64.11 & 52.84 & 66.72 \\
SpikeYOLO (OBB)$^\circ$ & 48.3M & 13.9\,+\,36.0 & 96.2 & 87.74 & 80.61 & 44.99 & 73.76 & 72.59 & 75.10 & 87.07 & 90.78 & 79.42 & 86.73 & 59.47 & 65.53 & 75.32 & 73.78 & 66.99 & 74.66\\
SpikeYOLO (Fused,OBB)$^\circ$ & 136.3M & 77.4 & 69.7 & 87.74 & 80.61 & 44.99 & 73.76 & 72.59 & 75.10 & 87.07 & 90.78 & 79.42 & 86.73 & 59.47 & 65.53 & 75.32 & 73.78 & 66.99 & 74.66\\
\hline
\rowcolor{gray!15} \textbf{SpikeAD (Ours)}$^\star$ & \textbf{46.5M} & \textbf{28.9} & \textbf{27.6} & 87.56 & 81.16 & 50.48 & 75.67 & 81.41 & 83.63 & 88.68 & 90.73 & 84.51 & 87.85 & 62.70 & 67.54 & 74.12 & 80.08 & 68.20 & \textbf{77.62} \\
\hline
\end{tabular}
\end{table*}
\begin{figure*}[t]
\centerline{\includegraphics[width=\linewidth]{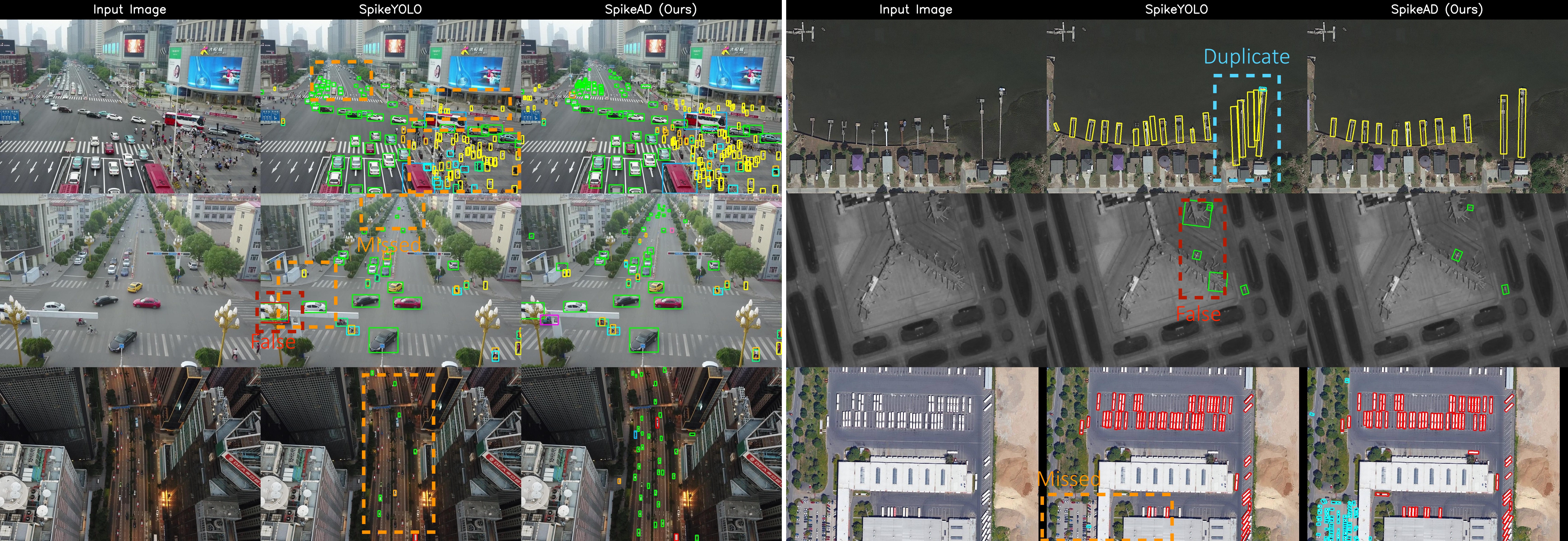}}\vspace{-0.2cm}
\caption{Sample prediction results on VisDrone and DOTA datasets and comparison with state-of-the-art SNN detector (SpikeYOLO). Left: VisDrone (HBB). Right: DOTA (OBB). SpikeYOLO does not support OBB and we kept the model by using our SpikeOBB detection head to train SpikeYOLO on DOTA for comparison.}
\label{fig:pred}
\end{figure*}

\subsubsection{Generalization to Oriented Object Detection in Remote Sensing (DOTA)}

To demonstrate the architectural flexibility and generalization capability of our approach, we evaluate SpikeAD on the DOTA-v1 remote sensing benchmark for oriented bounding box (OBB) detection. In UAV-based remote sensing scenarios, typically power constraints are lower. Therefore we report SpikeAD's M variant as the sole model in this task. As summarized in Table~\ref{tab:dota}, SpikeAD seamlessly adapts to OBB tasks by activating the spike-controlled orientation branch, achieving a highly competitive 77.62\% mAP. Compared to Transformer-based models (e.g., ARS-DETR, 73.79\% mAP) and heavy two-stage detectors (e.g., Oriented R-CNN, 75.87\% mAP), SpikeAD not only delivers superior detection accuracy but operates strictly within a low-power neuromorphic paradigm.SpikeAD achieves the highest DOTA accuracy among the evaluated SNN-based detectors, while remaining competitive with specialized ANN-based oriented detectors. S3Det-50~\cite{s3det}, which relies on ANN-to-SNN conversion, achieves only 66.72\% mAP while requiring a high parameter count of 121.1M and 93.6 mJ of energy. SpikeYOLO only supports HBB, therefore we also adapted SpikeYOLO for OBB detection by replacing the detection head with the proposed SpikeOBB head to serve as a direct baseline. While the unfused SpikeYOLO (OBB) achieves 74.66\% mAP, it remains a hybrid model; upon fusing it to enforce fully spike-driven execution (SpikeYOLO Fused, OBB), its parameter count explodes to 136.3M, consuming 69.7 mJ of energy. SpikeAD avoids this entirely. Utilizing the proposed SPC inference, SpikeAD achieves the highest accuracy (77.62\% mAP) with only 46.5M parameters and a remarkably low 27.6 mJ arithmetic inference energy (70.5\% and 60.4\% reduction compared to S3Det and fused SpikeYOLO (OBB), respectively). Notably, SpikeAD shows notable improvements on dense target clusters, e.g., Small Vehicles, 81.41\% AP (+8.82\%).

\subsubsection{Qualitative Visual Analysis}
To further validate these quantitative gains, Fig.~\ref{fig:pred} shows six sample detection results on both VisDrone (HBB, left) and DOTA (OBB, right). As visualized in the first and second image, SpikeYOLO misses numerous small objects located at far distances relative to the drone, whereas SpikeAD resolves these distant targets. Furthermore, SpikeYOLO has higher false positive predictions in complex urban environments (second image). In more challenging and high-altitude and darker scenarios (third image) where targets occupy a few pixels, SpikeAD's HRFPN and SGMS mechanism suggests better background noise suppression and enhances feature fusion, resulting in a substantially denser and more accurate prediction. Similar advantages are observed in remote sensing imagery: for dense and stretched objects, SpikeYOLO suffers from overlapping and duplicate predictions. It also falsely predicts structural jet bridges as planes (second image). Finally, similar to the observations on VisDrone, for high-altitude remote sensing, SpikeAD maintains better performance in detecting and correctly orienting tiny objects, suggesting that our neuromorphic architecture does not compromise on the complex spatial understanding required for OBB tasks. 


\subsection{Ablation Study}

\begin{figure*}[t]
\centerline{\includegraphics[width=\textwidth]{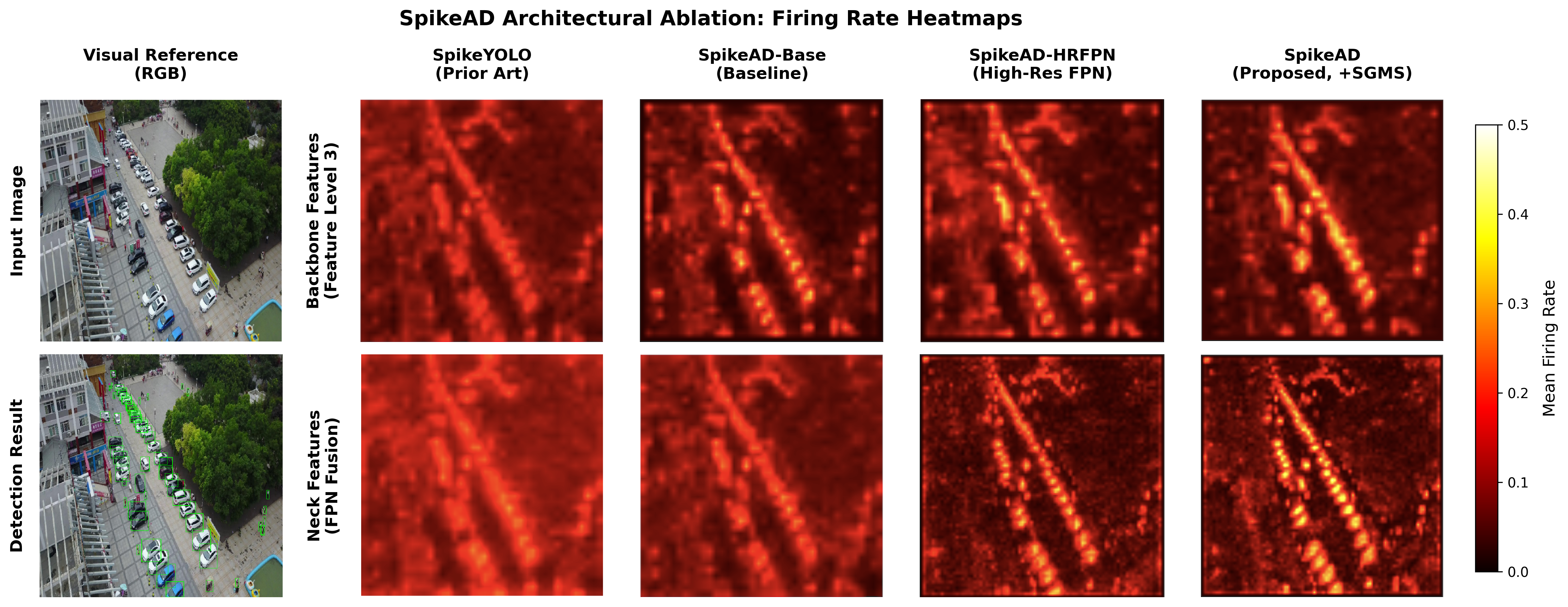}}\vspace{-0.4cm}
\caption{Ablation and explainability visualization via SNN firing heatmaps: evolution of spatial attention across architecture ablations. Global firing rate from left to right: 11.4\%, 9.7\%, 10.2\%, 11.1\%, 10.5\%.}
\label{fig:ablation}
\end{figure*}

To systematically validate the contributions of our proposed architectural components, we conduct a progressive ablation study on the VisDrone benchmark. Quantitative results across all variants are summarized in Table~\ref{tab:ablation} and qualitative evidence is depicted in Fig.~\ref{fig:ablation}.

\paragraph{Baseline Spiking Architecture} We designate SpikeAD-Base as the baseline model, which incorporates a standard three-scale feature pyramid~\cite{yolov10, yolov12, spikeyolo} with Spike MetaFormer blocks (Section~\ref{sec:spikeblocks}) using factorized spike mixers. Unlike SpikeYOLO's SepRepConv/RepConv spatial and channel mixers, which are either hybrid (MACs in unfused intermediate layers) or require dense-kernel expansion when fused, SpikeAD-Base has factorized mixers throughout the network, ensuring spike continuity at every sub-operator with no reparameterization overhead. Consequently, even the baseline achieves substantially lower energy (8.8 mJ at N-scale) while remaining fully spike-controlled. Visually, as demonstrated in the heatmaps (Fig.~\ref{fig:ablation}), ensuring spike continuity feature maps and replacing reparameterization-dependent spatial and channel mixers with our fully factorized spike mixers results in a visibly lower global firing rate and the baseline's attention is already much more densely concentrated on physical objects.

\begin{table}[t]
\centering
\caption{Progressive ablation on VisDrone. GACs are normalized with $\rho$.
  Baseline: 3-scale FPN with standard SSM and SFC\@.
  $+$HRFPN: high-resolution 4-scale bidirectional FPN.
  $+$MS: SGMS multi-scale dilated mixer without binary gate.
  $+$SGMS: complete SpikeAD, $^\star$spike-controlled.
  $\rho$: active spike rate;}
\label{tab:ablation}
\renewcommand{\arraystretch}{0.8}
\setlength{\tabcolsep}{2.2pt}
\small
\begin{tabular}{l|c c c c c c c}
\hline \textbf{Variant} & \textbf{Params} & \textbf{GACs} & $\boldsymbol{\rho}$ & \textbf{Energy} & $\boldsymbol{\eta}\uparrow$ & \textbf{mAP$_{50}$} & \textbf{mAP$_{50\text{-}95}$} \\ & \textbf{(M)} & \textbf{(G)} & \textbf{(\%)} & \textbf{(mJ)} & & & \\ \hline\hline
Baseline         & 13.2 & 9.3 & 27.2 &  8.8 & 17.8$\times$ & 36.2 & 20.8 \\
$+$HRFPN         & 14.5 & 12.4 & 28.5 & 12.0 & 16.8$\times$ & 38.8 & 22.9 \\
$+$MS            & 13.5 & 12.5 & 29.2 & 12.0 & 16.4$\times$ & 39.0 & 23.0 \\
\rowcolor{gray!20} $+$SGMS$^\star$ & \textbf{13.5} & \textbf{11.9} & \textbf{27.8} & \textbf{11.4} & \textbf{17.3}$\times$ & \textbf{39.3} & \textbf{23.2} \\
\hline
\multicolumn{8}{l}{{\normalfont\itshape Small}} \\
\hline
Baseline         & 23.1 & 15.9 & 26.6 & 15.1 & 18.3$\times$ & 38.3 & 22.2 \\
$+$HRFPN         & 25.5 & 20.9 & 27.8 & 19.9 & 17.4$\times$ & 41.3 & 24.4 \\
$+$MS            & 23.4 & 20.4 & 27.7 & 19.5 & 17.4$\times$ & 41.8 & 24.7 \\
\rowcolor{gray!20} $+$SGMS$^\star$ & \textbf{23.4} & \textbf{20.2} & \textbf{27.4} & \textbf{19.3} & \textbf{17.6}$\times$ & \textbf{42.5} & \textbf{25.2} \\
\hline
\multicolumn{8}{l}{{\normalfont\itshape Medium}} \\
\hline
Baseline         & 47.7 & 32.6 & 26.2 & 30.7 & 18.7$\times$ & 41.5 & 24.6 \\
$+$HRFPN         & 51.0 & 39.7 & 27.0 & 37.5 & 18.0$\times$ & 44.9 & 27.1 \\
$+$MS            & 46.2 & 39.7 & 27.7 & 37.8 & 17.5$\times$ & 45.0 & 27.1 \\
\rowcolor{gray!20} $+$SGMS$^\star$ & \textbf{46.2} & \textbf{36.5} & \textbf{25.5} & \textbf{34.8} & \textbf{18.9}$\times$ & \textbf{46.2} & \textbf{27.9} \\
\hline
\end{tabular}
\vspace{-0.3cm}
\end{table}

\paragraph{High-Resolution FPN (HRFPN)} Redesigning the neck to a four-scale high-resolution bidirectional feature pyramid tailored for small object detection in aerial imagery yields a substantial +2.6\% to +3.4\% mAP@50 improvement across variants. This quantitative leap is visualized in Fig.~\ref{fig:ablation}: at the neck level, HRFPN demonstrates substantially sharper multi-scale feature aggregation on the target objects compared to the baseline neck. Nevertheless, this additional scale naturally increases the parameter count and slightly elevates the active spike rate $\rho$.

\paragraph{Multi-Scale Extraction and Binary Gating} We replace the standard SSM with the proposed SGMS block. We also decouple SGMS to isolate the effects of its multi-scale dilated parallel branches and its binary channel gate. Adding only the multi-scale branches (+MS) slightly improves mAP but increases the active spike rate $\rho$. This indicates that broader receptive fields without activity-driven selection marginally raise the active spike fraction and leave the energy reduction factor $\eta$ unchanged or worse. Multi-scale spatial context alone is therefore insufficient; the gating mechanism is necessary to translate it into both accuracy and efficiency gains. Integrating it with binary channel gate (the complete +SGMS configuration in table~\ref{tab:ablation}) resolves this by dynamically suppressing weak spike channels. As depicted in the final column of Fig.~\ref{fig:ablation}, SGMS effectively filters out the residual environmental noise and the refined backbone feature representation cascades into the neck, producing the highest and most precise firing concentration on the targets. Consequently, SGMS not only maximizes detection accuracy (reaching 39.3\%, 42.5\%, 46.2\% mAP@50 across the three variants) but simultaneously reduces the total parameters and suppresses $\rho$.

\paragraph{The Compounding Efficiency Effect} Ultimately, the ablation study reveals an effect that SGMS ensures that the network scales sub-linearly in computational cost. For instance, SpikeAD-M achieves an arithmetic energy reduction factor $\eta=18.9\times$ compared to an equivalent ANN, which is even higher than the simpler 3-scale baseline (18.7$\times$) despite an extra pyramid scale and a richer backbone spatial mixer, indicating that the reduction in active-spike rate associated with SGMS compensates for part of the additional operation budget introduced by the four-scale pyramid. Under the adopted arithmetic-energy model, the N and M variants require 11.4 mJ and 34.8 mJ, respectively, keeping all variants comfortably below 35 mJ under the adopted arithmetic energy model for power-constrained edge UAV platforms. 


\section{Hardware Realizability}
\label{sec:hardware}

\begin{figure*}[t]
\centerline{\includegraphics[width=\linewidth]{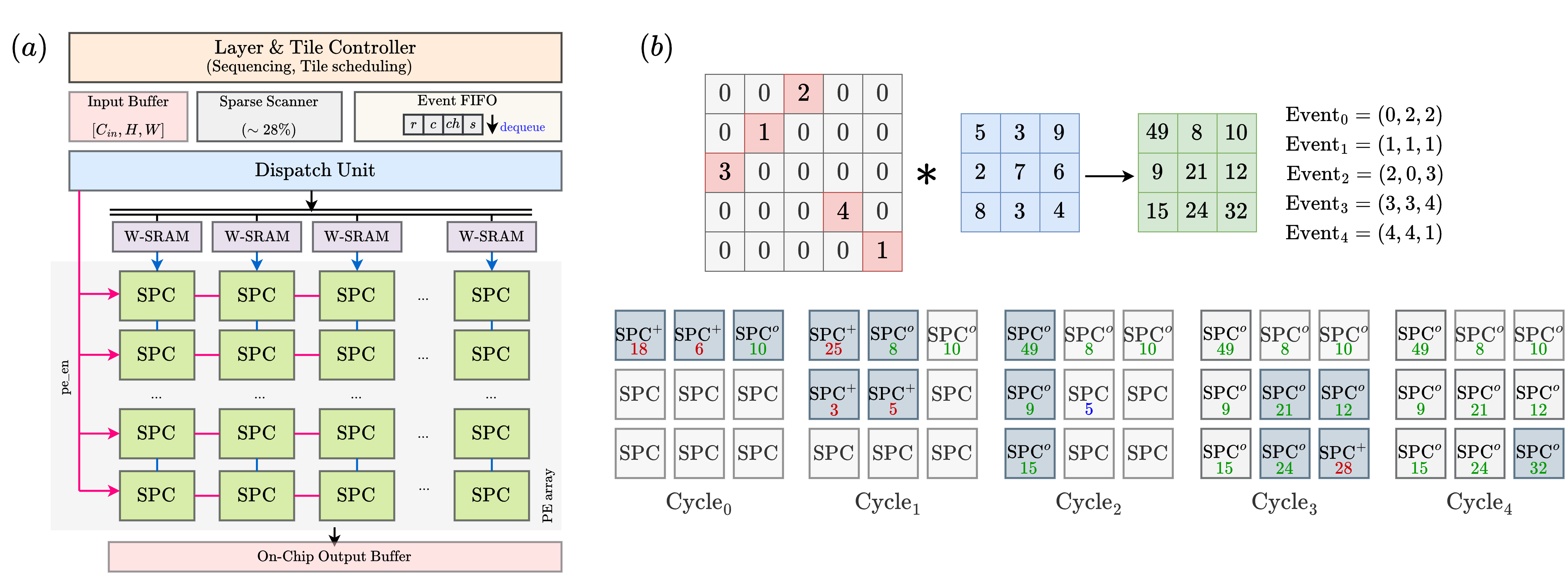}}\vspace{-0.4cm}
\caption{Illustrative custom chip design mapping: (a) Conceptual architecture of the Sparse Spike-Controlled Accelerator (SSCA) with an optional nonzero-event extractor and FIFO routing spikes to the SPC array, analogous to GPU scheduling but optimized for spike events; only the individual SPC primitive is implemented in RTL in this work; (b) Execution dataflow example for a $\operatorname{SC}_{DW,3}$ on a $5\times5$ feature map. Only five nonzero input events are dispatched, whereas a dense formulation evaluates all 25 input positions. Actual cycle counts depend on array parallelism, memory organization, fan-out, and scheduling. }
\label{fig:arch_dflow}
\end{figure*}

SpikeAD is primarily an algorithmic architecture designed under hardware-aware constraints, including spike activations, bounded integer spike levels, BN folding, binary channel gating. The SPC RTL implementation validates the multiplier-free realizability of its central arithmetic operator;
The Sparse Spike-Controlled Accelerator (SSCA) (Fig.~\ref{fig:arch_dflow}) is an illustrative mapping to present a scalable architecture with inference abstraction, while complete accelerator implementation and system-level measurements are outside the scope of this work.

The measured active-spike rates indicate that more than 72\% of dense-equivalent spike events are zero and can be skipped by the proposed mapping. However, current AI accelerators (GPUs, TPUs, mobile NPUs) are optimized for dense matrix multiplication and cannot natively exploit this dynamic sparsity as existing structured-sparsity mechanisms, such as NVIDIA’s 2:4 format primarily target fixed, hardware-compatible sparsity patterns in weights and software deployment stacks (e.g. ONNX, TensorRT) do not natively support the unstructured, dynamic sparsity of SNNs. Conversely, existing neuromorphic chips~\cite{loihi} have their execution model designed around neuromorphic neuron and synaptic operations rather than the specific multi-level spike-controlled arithmetic blocks employed by SpikeAD, and have not been demonstrated on small object detection at this scale. This motivates a custom hardware abstraction to realize SpikeAD's execution semantics.

\begin{figure}[t]
\centerline{\includegraphics[width=\linewidth]{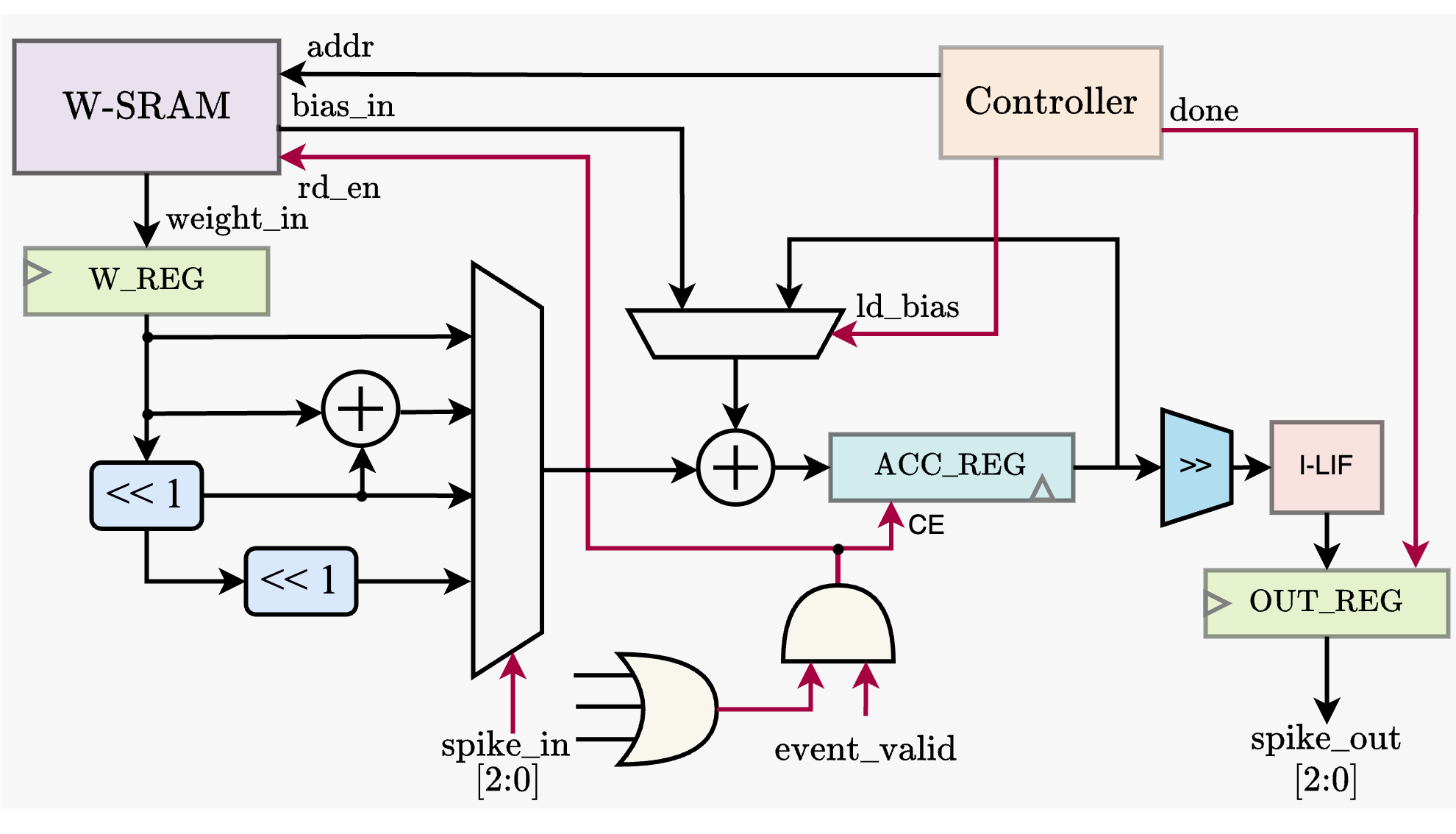}}
\caption{Spike Processing Core (SPC) microarchitecture and datapath with 3-stage pipeline. The spike value acts as a 3-bit control signal selecting a shifted weight into the accumulator via combinatorial logic. Zero spikes disable the datapath via clock-enable gating. The W-SRAM output register is shown for pipeline clarity but lies outside the synthesized SPC boundary. No multiplier circuit is required.}
\label{fig:spc}
\vspace{-0.5cm}
\end{figure}

\subsection{Spike Processing Core (SPC) Microarchitecture}
\label{sec:spc}

To physicalize the spike-controlled convolution paradigm defined in Section~\ref{sec:methodology}, unlike conventional MAC unit where both weight and activation are arithmetic inputs to a multiplier, we propose the Spike Processing Core (SPC), which exploits multi-level spikes exclusively as a control signal. SpikeAD retains discrete spike-mediated computation, while zero-valued activations enable compute gating in the proposed SPC mapping. As shown in Fig.~\ref{fig:spc}, the spike value $s \in \{0,\ldots,4\}$ acts as a 3-bit selector that routes a pre-shifted weight into the accumulator via purely combinatorial logic. For $s=0$ the clock-enable is deasserted and the datapath is idle. The $s=3$ case requires one lightweight combinatorial pre-add before the single accumulation; $K{=}4$ is the largest bound for which the non-power-of-two cases reduce to the single value $s{=}3$. $s = \{1,2,4\}$ requires only direct routing or zero-cost wiring shifts. The selected value enters a 32-bit accumulator via a single adder. After all active events have been processed, a pipeline register captures the result and a fixed-point right-shift feeds four comparators that implement the I-LIF threshold to produce the output spike $s'$. In the illustrative SSCA, zero events can be removed before weight-memory requests, while the registered SRAM output is delivered to the SPC together with the aligned spike value. Although it dispatches only nonzero events, support for $s=0$ enables local clock gating and allows the SPC to operate with either sparse-event or dense-stream front ends. The illustrative SSCA adopts a weight-stationary mapping; its system-level data-movement benefits require full accelerator evaluation.

\begin{table}[t]
\centering
\caption{VisDrone performance under FP32 vs. W8A3 quantization. W8A3 energy relies on INT32 accumulation (0.1 pJ/add).} \vspace{-0.2cm}
\label{tab:quantization}
\renewcommand{\arraystretch}{0.6}
\setlength{\tabcolsep}{6pt}
\footnotesize
\begin{tabular}{ll ccc}
\toprule
\textbf{Variant} & \textbf{Format} & \textbf{mAP$_{50}$} & \textbf{mAP$_{50\text{-}95}$} & \textbf{Energy (mJ)} \\
\midrule
\multirow{2}{*}{SpikeAD-N} 
 & FP32 & 39.3 & 23.2 & 11.4 \\
 & W8A3 & 39.1 & 23.0 & 1.27 \\
\midrule
\multirow{2}{*}{SpikeAD-S} 
 & FP32 & 42.5 & 25.2 & 19.3 \\
 & W8A3 & 42.3 & 25.1 & 2.14 \\
\midrule
\multirow{2}{*}{SpikeAD-M} 
 & FP32 & 46.2 & 27.9 & 34.8 \\
 & W8A3 & 46.0 & 27.8 & 3.87 \\
\bottomrule
\end{tabular}
\vspace{-0.5cm}
\end{table}

\subsection{Quantization}
\label{sec:quant}

While Section~\ref{sec:results} adopts the Horowitz FP32 model for fair algorithmic benchmarking, dedicated neuromorphic hardware natively operates on fixed-point logic. SpikeAD bounds activations to discrete levels ($s \in [0, 4]$, 3-bit), enabling post-training INT8 per-channel symmetric uniform weight quantization (W8A3) without activation calibration. As shown in Table \ref{tab:quantization}, this yields negligible accuracy degradation on VisDrone. Crucially, mapping the accumulation to INT32 reduces the addition energy from 0.9 pJ (FP32) to 0.1 pJ. Consequently, the arithmetic inference energy of SpikeAD-M drops to 3.87 mJ, further highlighting the architectural advantage of multiplier-free execution for ultra-low-power edge computing. For a quantized ANN comparison, a conventional INT8 MAC is 0.3 pJ~\cite{horowitz}, under this accounting the dense-equivalent operations of SpikeAD-M would require 42.9 mJ, and the W8A3 shift-add datapath retains an 11.1× reduction. Compared to ANNs, SpikeAD-S can match EDNet-S at 2.6x lower arithmetic energy, and SpikeAD-M exceeds YOLOv12-M by 2.9 points at 2.8× lower energy while remaining within 0.9 points of EDNet-M at 3.4× lower energy.

\subsection{RTL Synthesis}

We implement the SPC in synthesizable Verilog RTL and perform place-and-route on the Xilinx ZCU104 FPGA (XCZU7EV, 230,400 LUTs / 460,800 FFs). Post-implementation utilization per SPC is 93 LUTs, 36 flip-flops, and zero DSP48 blocks targeting the quantized inference. The absence of DSP resources is architecturally important: FPGA MAC-based accelerators are typically DSP-limited (ZCU104 provides 1,728 DSP48 blocks). The low per-core resource utilization suggests the potential for substantial core replication on a single FPGA. However, practical array size will also depend on weight-memory capacity, routing congestion, fan-out, bandwidth, and controller overhead. The complete hardware design and analysis is outside the scope of this work.

\subsection{Illustrative Accelerator Mapping}
\label{sec:system}

The SPC is embedded within a scalable, event-driven system architecture, serving as the fundamental compute primitive for our proposed Sparse Spike-Controlled Accelerator (SSCA). As illustrated in Fig.~\ref{fig:arch_dflow}(a), an optional nonzero-event extractor may scan the 3-bit feature map from on-chip memory (BRAM on FPGA, SRAM on ASIC and emit event tuples $(r, c, \text{ch}, s)$ for $s\neq 0$, reducing traffic to the dispatch unit and SPC array. By discarding up to 74\% of the inputs at the memory boundary, no compute logic is wasted on zero-valued activations. Emitted events would be buffered in a hardware FIFO, which decouples the scan rate from the SPC array consumption rate and acts as a natural burst buffer for the irregular, activity-dependent event stream. Events would be dequeued one per cycle and processed by a dispatch unit. For each event at input position $(r, c)$, the set of affected output positions $(r', c')$ within the kernel window (at most $k^2$ positions), and broadcasts (fan-out) the event alongside the corresponding weight address to the matching SPCs. A tile controller would manage the lifecycle: it would load bias values at the start of each output position, track the completion state of each output accumulation, and assert the \texttt{done} signal to capture results, and advance the tile pointer to the next spatial or channel batch. 

This dispatch-execute paradigm shares a deep structural analogy with modern GPU architectures. However, instead of a central scheduler dispatching dense instruction streams (warps) to an array of CUDA cores, the SSCA dispatch unit would route sparse spike events to an array of SPCs. Fig.~\ref{fig:arch_dflow}(b) shows an example on the theoretical execution dataflow.

The SPC count is a synthesis-time parameter. An $N \times M$ array would assign $N$ rows to output spatial positions within the current tile and $M$ columns to output channels, each with dedicated local weight memory. The output map is processed in $\lceil H_\text{out} W_\text{out} / N \rceil \times \lceil C_\text{out} / M \rceil$ tile passes. A single dispatched event could activate up to $k^2 \times M$ SPCs concurrently, amortizing dispatch overhead across a large number of parallel accumulations. 

The channel gate (Eq.~\ref{eq:channel_gate}) requires spatial accumulation before the grouped pointwise projection. This dependency does not constrain the mutually independent dilated branches and can be accommodated within the tile-sequential dataflow. Computing the gate requires $CH'W'$ spatial accumulation operations, whereas the subsequent projection requires approximately $C^2H'W'/g$ spike-controlled operations. Its relative arithmetic overhead therefore scales approximately as $g/C$, excluding buffering and control. The channel-wise sums can be accumulated while $S_{MS}$ is written to the intermediate feature buffer.  Once all spatial tiles have been processed, the $C$ gate values are resolved before the projection begins, requiring $C$ additional words. Suppressed channels can then be excluded from projection dispatch, weight-memory access, and accumulation. Because the gate is resolved once per block rather than per event, its control overhead is amortized across the subsequent projection.

While cycle-accurate latency characterization of a full SSCA array depends on implementation-specific variables (e.g., memory bandwidth, tile scheduling, and target process node), the SPC avoids multiplier logic and is expected to allow a shorter critical path. Combined with the ${\approx}72\%$ zero-event suppression which reduces the number of events presented to the dispatch stage, SpikeAD is well-suited for latency-optimized neuromorphic accelerator designs.

The SSCA in Fig.~\ref{fig:arch_dflow} is therefore a proposed system architecture rather than a complete implemented accelerator. The RTL results validate the SPC arithmetic primitive, while end-to-end throughput, latency, memory traffic, power, and energy remain outside the scope of this study. Ultimately, the SPC and this system architecture constitute the hardware vision for SpikeAD: an algorithmic framework co-designed with a class of reconfigurable, scalable dedicated accelerators that can fully exploit multiplication-free and spike-sparse computation for autonomous aerial platforms.
\section{Conclusion and Future Work}

In this paper, we introduced SpikeAD, a directly-trained, fully spike-controlled neuromorphic architecture tailored for highly energy-constrained aerial object detection. By replacing standard dense MAC operations with a shift-add Spike Processing Core (SPC) primitive, and introducing the Spike-Gated Multi-Scale (SGMS) block for adaptive, activity-driven feature extraction, SpikeAD demonstrates that bounded integer spikes can be used as control operands for multiplier-free shift-add inference. Evaluated on the VisDrone and DOTA benchmarks, SpikeAD improves the accuracy-efficiency trade-off relative to the evaluated SNN baselines, matching the detection precision of specialized conventional models while achieving substantial reductions in estimated arithmetic energy. Supported by RTL synthesis of its central compute primitive, SpikeAD establishes a multiplier-free, hardware-aware algorithmic foundation for the next generation of ultra-low-power scalable neuromorphic accelerators for edge AI in aerial use cases. Future research will focus on two primary directions. First, we aim to transition the proposed RTL primitive to a full custom hardware accelerator to evaluate system-level energy metrics and speed. Second, because SpikeAD fundamentally retains a 5D tensor representation throughout its architecture, future work will study neuromorphic data from native event cameras during dynamic UAV flights.

\bibliographystyle{ieeetr}
\bibliography{ref}

@article{nature,
  author    = {Kaushik Roy and Akhilesh Jaiswal and Priyadarshini Panda},
  title     = {{Towards Spike-Based Machine Intelligence with Neuromorphic Computing}},
  journal   = {Nature},
  year      = {2019},
  volume     = {575},
  number     = {7784},
  pages      = {607--617},
  month      = nov,
  doi        = {10.1038/s41586-019-1677-2},
  url        = {https://doi.org/10.1038/s41586-019-1677-2},
  issn       = {1476-4687}
}

@article{ne,
  author    = {Carver Mead},
  title     = {{How We Created Neuromorphic Engineering}},
  journal   = {Nature Electronics},
  year      = {2020},
  volume    = {3},
  number    = {7},
  pages     = {434--435},
  month     = jul,
  doi       = {10.1038/s41928-020-0448-2},
  url       = {https://doi.org/10.1038/s41928-020-0448-2}
}

@article{ncs,
  author  = {Catherine D. Schuman and Shruti R. Kulkarni and Maryam Parsa and J. Parker Mitchell and Prasanna Date and Bill Kay},
  title   = {{Opportunities for Neuromorphic Computing Algorithms and Applications}},
  journal = {Nature Computational Science},
  year    = {2022},
  volume  = {2},
  number  = {1},
  pages   = {10--19},
  month   = jan,
  doi     = {10.1038/s43588-021-00184-y},
  url     = {https://doi.org/10.1038/s43588-021-00184-y},
  issn    = {2662-8457}
}

@ARTICLE{loihi,
  author={Davies, Mike and Srinivasa, Narayan and Lin, Tsung-Han and Chinya, Gautham and Cao, Yongqiang and Choday, Sri Harsha and Dimou, Georgios and Joshi, Prasad and Imam, Nabil and Jain, Shweta and Liao, Yuyun and Lin, Chit-Kwan and Lines, Andrew and Liu, Ruokun and Mathaikutty, Deepak and McCoy, Steven and Paul, Arnab and Tse, Jonathan and Venkataramanan, Guruguhanathan and Weng, Yi-Hsin and Wild, Andreas and Yang, Yoonseok and Wang, Hong},
  journal={IEEE Micro}, 
  title={{Loihi: A Neuromorphic Manycore Processor with On-Chip Learning}}, 
  year={2018},
  volume={38},
  number={1},
  pages={82-99},
  doi={10.1109/MM.2018.112130359}}

@ARTICLE{heteroaisoc,
  author={Song, Zhifan and Stratigopoulos, Haralampos-G. and Aboushady, Hassan},
  journal={IEEE Transactions on Very Large Scale Integration (VLSI) Systems}, 
  title={{A Heterogeneous Neural Network Accelerator for End-to-End Multitask RF Signal Recognition}}, 
  year={2026},
  volume={},
  number={},
  pages={1-14},
  doi={10.1109/TVLSI.2026.3715734}}

@article{tianjic,
  author    = {J. Pei and L. Deng and S. Song and M. Zhao and Y. Zhang and S. Wu and G. Wang and Z. Zou and Z. Wu and W. He and F. Chen and N. Deng and S. Chen and T. Wu and R. A. Xie and L. Han and Z. Zhang and T. L. Zheng and J. Hao and F. Shen and G. Wang and Y. Min and Y. Wang and Z. Zhu and Y. Xiao and S. Wu and J. Chen and H. Wu and L. Shen and D. Li and X. Hu and R. Xu and Y. Zhou and X. Ma and T. Shi and Y. Wang and H. Lin and F. Lv and G. Ji and X. Han and Z. Li and Y. Wang and Y. Liu and C. Liang and J. Miao and J. Peng and C. Zhao and Y. Wang and H. Fang and W. Yu and C. Zhou and G. Dai and Y. Wang and Y. Wu and H. Luo and Y. Xie and L. Shi},
  title     = {{Towards Artificial General Intelligence with Hybrid Tianjic Chip Architecture}},
  journal   = {Nature},
  volume    = {572},
  number    = {7767},
  pages     = {106--111},
  year      = {2019},
  month     = aug,
  doi       = {10.1038/s41586-019-1424-8},
  url       = {https://doi.org/10.1038/s41586-019-1424-8}
}

@inproceedings{towards,
author = {Seekings, James and Chandarana, Peyton and Ardakani, Mahsa and Mohammadi, MohammadReza and Zand, Ramtin},
title = {{Towards Efficient Deployment of Hybrid SNNs on Neuromorphic and Edge AI Hardware}},
year = {2025},
isbn = {9798350368659},
publisher = {IEEE Press},
url = {https://doi.org/10.1109/ICONS62911.2024.00018},
doi = {10.1109/ICONS62911.2024.00018},
booktitle = {Proceedings of the International Conference on Neuromorphic Systems},
pages = {71–77},
numpages = {7},
location = {Arlington, Virginia, USA},
series = {ICONS '24}
}

@article{lif,
title = {{Networks of spiking neurons: The third generation of neural network models}},
journal = {Neural Networks},
volume = {10},
number = {9},
pages = {1659-1671},
year = {1997},
issn = {0893-6080},
doi = {https://doi.org/10.1016/S0893-6080(97)00011-7},
url = {https://www.sciencedirect.com/science/article/pii/S0893608097000117},
author = {Wolfgang Maass},
}

@ARTICLE{surrogate,
  author={Neftci, Emre O. and Mostafa, Hesham and Zenke, Friedemann},
  journal={IEEE Signal Processing Magazine}, 
  title={{Surrogate Gradient Learning in Spiking Neural Networks: Bringing the Power of Gradient-Based Optimization to Spiking Neural Networks}}, 
  year={2019},
  volume={36},
  number={6},
  pages={51-63},
  doi={10.1109/MSP.2019.2931595}}

@INPROCEEDINGS{biodrone,
  author={Lamberti, Lorenzo and Bompani, Luca and Kartsch, Victor Javier and Rusci, Manuele and Palossi, Daniele and Benini, Luca},
  booktitle={2023 Design, Automation \& Test in Europe Conference \& Exhibition (DATE)}, 
  title={{Bio-inspired Autonomous Exploration Policies with CNN-based Object Detection on Nano-drones}}, 
  year={2023},
  volume={},
  number={},
  pages={1-6},
  doi={10.23919/DATE56975.2023.10137154}}

@InProceedings{coco,
author="Lin, Tsung-Yi
and Maire, Michael
and Belongie, Serge
and Hays, James
and Perona, Pietro
and Ramanan, Deva
and Doll{\'a}r, Piotr
and Zitnick, C. Lawrence",
editor="Fleet, David
and Pajdla, Tomas
and Schiele, Bernt
and Tuytelaars, Tinne",
title="Microsoft COCO: Common Objects in Context",
booktitle="Computer Vision -- ECCV 2014",
year="2014",
publisher="Springer International Publishing",
address="Cham",
pages="740--755",
isbn="978-3-319-10602-1"
}

@ARTICLE{visdrone,
  author={Zhu, Pengfei and Wen, Longyin and Du, Dawei and Bian, Xiao and Fan, Heng and Hu, Qinghua and Ling, Haibin},
  journal={IEEE Transactions on Pattern Analysis and Machine Intelligence}, 
  title={{Detection and Tracking Meet Drones Challenge}}, 
  year={2022},
  volume={44},
  number={11},
  pages={7380-7399},
  doi={10.1109/TPAMI.2021.3119563}}

@ARTICLE{dota,
  author={Ding, Jian and Xue, Nan and Xia, Gui-Song and Bai, Xiang and Yang, Wen and Yang, Michael and Belongie, Serge and Luo, Jiebo and Datcu, Mihai and Pelillo, Marcello and Zhang, Liangpei},
  journal={IEEE Transactions on Pattern Analysis and Machine Intelligence},
  title={{Object Detection in Aerial Images: A Large-Scale Benchmark and Challenges}},
  year={2021},
  volume={},
  number={},
  pages={1-1},
  doi={10.1109/TPAMI.2021.3117983}}

@InProceedings{yolo,
    author = {Redmon, Joseph and Divvala, Santosh and Girshick, Ross and Farhadi, Ali},
    title = {{You Only Look Once: Unified, Real-Time Object Detection}},
    booktitle = {Proceedings of the IEEE Conference on Computer Vision and Pattern Recognition (CVPR)},
    month = {June},
    year = {2016}
}

@article{ultralytics,
  author  = {Jocher, Glenn and Qiu, Jing and Liu, Mengyu and Lyu, Shuai and Akyon, Fatih Cagatay and Kalfaoglu, Muhammet Esat},
  title   = {{Ultralytics YOLO26: Unified Real-Time End-to-End Vision Models}},
  journal = {arXiv preprint arXiv:2606.03748},
  year    = {2026},
  doi     = {10.48550/arXiv.2606.03748}
}

@inproceedings{yolov10,
 author = {Wang, Ao and Chen, Hui and Liu, Lihao and Chen, Kai and Lin, Zijia and Han, Jungong and Ding, Guiguang},
 booktitle = {Advances in Neural Information Processing Systems},
 doi = {10.52202/079017-3429},
 editor = {A. Globerson and L. Mackey and D. Belgrave and A. Fan and U. Paquet and J. Tomczak and C. Zhang},
 pages = {107984--108011},
 publisher = {Curran Associates, Inc.},
 title = {{YOLOv10: Real-Time End-to-End Object Detection}},
 url = {https://proceedings.neurips.cc/paper_files/paper/2024/file/c34ddd05eb089991f06f3c5dc36836e0-Paper-Conference.pdf},
 volume = {37},
 year = {2024}
}

@inproceedings{yolov12,
 author = {Tian, Yunjie and Ye, Qixiang and DOERMANN, DAVID},
 booktitle = {Advances in Neural Information Processing Systems},
 editor = {D. Belgrave and C. Zhang and H. Lin and R. Pascanu and P. Koniusz and M. Ghassemi and N. Chen},
 pages = {78433--78457},
 publisher = {Curran Associates, Inc.},
 title = {{YOLOv12: Attention-Centric Real-Time Object Detectors}},
 url = {https://proceedings.neurips.cc/paper_files/paper/2025/file/7103444259031cc58051f8c9a4868533-Paper-Conference.pdf},
 volume = {38},
 year = {2025}
}

@ARTICLE{fasterrcnn,
  author={Ren, Shaoqing and He, Kaiming and Girshick, Ross and Sun, Jian},
  journal={IEEE Transactions on Pattern Analysis and Machine Intelligence}, 
  title={{Faster R-CNN: Towards Real-Time Object Detection with Region Proposal Networks}}, 
  year={2017},
  volume={39},
  number={6},
  pages={1137-1149},
  doi={10.1109/TPAMI.2016.2577031}}

@inproceedings{dino,
  title={{DINO: DETR with Improved DeNoising Anchor Boxes for End-to-End Object Detection}},
  author={Zhang, Hao and Li, Feng and Liu, Shilong and Zhang, Lei and Su, Hang and Zhu, Jun and Ni, Lionel M. and Shum, Heung-Yeung},
  booktitle={International Conference on Learning Representations (ICLR)},
  url={https://openreview.net/forum?id=3mRwyG5one},
  year={2023}
}

@InProceedings{rtdetr,
    author    = {Zhao, Yian and Lv, Wenyu and Xu, Shangliang and Wei, Jinman and Wang, Guanzhong and Dang, Qingqing and Liu, Yi and Chen, Jie},
    title     = {{DETRs Beat YOLOs on Real-time Object Detection}},
    booktitle = {Proceedings of the IEEE/CVF Conference on Computer Vision and Pattern Recognition (CVPR)},
    month     = {June},
    year      = {2024},
    pages     = {16965-16974}
}

@ARTICLE{vamyolox,
  author={Yang, Yahu and Gao, Xiangzhou and Wang, Yu and Song, Shenmin},
  journal={IEEE Sensors Journal}, 
  title={{VAMYOLOX: An Accurate and Efficient Object Detection Algorithm Based on Visual Attention Mechanism for UAV Optical Sensors}}, 
  year={2023},
  volume={23},
  number={11},
  pages={11139-11155},
  doi={10.1109/JSEN.2022.3219199}}

@INPROCEEDINGS{edgeyolo,
  author={Liu, Shihan and Zha, Junlin and Sun, Jian and Li, Zhuo and Wang, Gang},
  booktitle={2023 42nd Chinese Control Conference (CCC)}, 
  title={{EdgeYOLO: An Edge-Real-Time Object Detector}}, 
  year={2023},
  volume={},
  number={},
  pages={7507-7512},
  doi={10.23919/CCC58697.2023.10239786}}

@INPROCEEDINGS{ednet,
  author={Song, Zhifan and Zhang, Yuan and Ebayyeh, Abd Al Rahman M. Abu},
  booktitle={21st IEEE International Conference on Ubiquitous Intelligence and Computing - 2024 IEEE Smart World Congress (SWC)}, 
  title={{EDNet: Edge-Optimized Small Target Detection in UAV Imagery - Faster Context Attention, Better Feature Fusion, and Hardware Acceleration}}, 
  year={2024},
  volume={},
  number={},
  pages={829-838},
  doi={10.1109/SWC62898.2024.00141}}

@INPROCEEDINGS{cslyolo,
  author={Zhang, Yu-Ming and Lee, Chun-Chieh and Hsieh, Jun-Wei and Fan, Kuo-Chin},
  booktitle={2022 IEEE International Symposium on Circuits and Systems (ISCAS)}, 
  title={{CSL-YOLO: A Cross-Stage Lightweight Object Detector with Low FLOPs}}, 
  year={2022},
  volume={},
  number={},
  pages={2730-2734},
  doi={10.1109/ISCAS48785.2022.9937880}}

@inproceedings{spikingyolo,
  author    = {Seijoon Kim and Seongsik Park and Byunggook Na and Sungroh Yoon},
  title     = {{Spiking-YOLO: Spiking Neural Network for Energy-Efficient Object Detection}},
  booktitle = {Proceedings of the AAAI Conference on Artificial Intelligence},
  pages      = {11270--11277},
  year       = {2020},
  doi        = {10.1609/aaai.v34i07.6787}
}

@InProceedings{emsyolo,
    author    = {Su, Qiaoyi and Chou, Yuhong and Hu, Yifan and Li, Jianing and Mei, Shijie and Zhang, Ziyang and Li, Guoqi},
    title     = {{Deep Directly-Trained Spiking Neural Networks for Object Detection}},
    booktitle = {Proceedings of the IEEE/CVF International Conference on Computer Vision (ICCV)},
    month     = {October},
    year      = {2023},
    pages     = {6555-6565}
}

@ARTICLE{fpgayolo1,
  author={Nguyen, Duy Thanh and Nguyen, Tuan Nghia and Kim, Hyun and Lee, Hyuk-Jae},
  journal={IEEE Transactions on Very Large Scale Integration (VLSI) Systems}, 
  title={{A High-Throughput and Power-Efficient FPGA Implementation of YOLO CNN for Object Detection}}, 
  year={2019},
  volume={27},
  number={8},
  pages={1861-1873},
  doi={10.1109/TVLSI.2019.2905242}}

@INPROCEEDINGS{fpgayolo2,
  author={Zhang, Jinming and Cheng, Lifu and Li, Cen and Li, Yongfu and He, Guanghui and Xu, Ningyi and Lian, Yong},
  booktitle={2021 IEEE International Symposium on Circuits and Systems (ISCAS)}, 
  title={{A Low-Latency FPGA Implementation for Real-Time Object Detection}}, 
  year={2021},
  volume={},
  number={},
  pages={1-5},
  doi={10.1109/ISCAS51556.2021.9401577}}

@INPROCEEDINGS{poolformer,
  author={Yu, Weihao and Luo, Mi and Zhou, Pan and Si, Chenyang and Zhou, Yichen and Wang, Xinchao and Feng, Jiashi and Yan, Shuicheng},
  booktitle={2022 IEEE/CVF Conference on Computer Vision and Pattern Recognition (CVPR)}, 
  title={{MetaFormer is Actually What You Need for Vision}}, 
  year={2022},
  volume={},
  number={},
  pages={10809-10819},
  doi={10.1109/CVPR52688.2022.01055}}

@inproceedings{metaspikeformer,
 author = {Yao, Man and Hu, Jiakui and Hu, Tianxiang and Xu, Yifan and Zhou, Zhaokun and Tian, Yonghong and XU, Bo and Li, Guoqi},
 booktitle = {International Conference on Learning Representations},
 editor = {B. Kim and Y. Yue and S. Chaudhuri and K. Fragkiadaki and M. Khan and Y. Sun},
 pages = {52885--52907},
 title = {{Spike-driven Transformer V2: Meta Spiking Neural Network Architecture Inspiring the Design of Next-generation Neuromorphic Chips}},
 volume = {2024},
 year = {2024}
}

@ARTICLE{espikeformer,
  author={Yao, Man and Qiu, Xuerui and Hu, Tianxiang and Hu, Jiakui and Chou, Yuhong and Tian, Keyu and Liao, Jianxing and Leng, Luziwei and Xu, Bo and Li, Guoqi},
  journal={IEEE Transactions on Pattern Analysis and Machine Intelligence}, 
  title={{Scaling Spike-Driven Transformer With Efficient Spike Firing Approximation Training}}, 
  year={2025},
  volume={47},
  number={4},
  pages={2973-2990},
  doi={10.1109/TPAMI.2025.3530246}}

@inproceedings{spikeyolo,
  author    = {Xinhao Luo and Man Yao and Yuhong Chou and Bo Xu and Guoqi Li},
  title     = {{Integer-Valued Training and Spike-Driven Inference Spiking Neural Network for High-Performance and Energy-Efficient Object Detection}},
  booktitle = {Computer Vision -- ECCV 2024},
  publisher = {Springer Nature Switzerland},
  address   = {Cham},
  pages     = {253--272},
  year      = {2025},
  isbn      = {978-3-031-73411-3}
}

@InProceedings{hybridspikingvit,
  title = 	 {{Hybrid Spiking Vision Transformer for Object Detection with Event Cameras}},
  author =       {Xu, Qi and Deng, Jie and Shen, Jiangrong and Chen, Biwu and Tang, Huajin and Pan, Gang},
  booktitle = 	 {Proceedings of the 42nd International Conference on Machine Learning},
  pages = 	 {69147--69159},
  year = 	 {2025},
  editor = 	 {Singh, Aarti and Fazel, Maryam and Hsu, Daniel and Lacoste-Julien, Simon and Berkenkamp, Felix and Maharaj, Tegan and Wagstaff, Kiri and Zhu, Jerry},
  volume = 	 {267},
  series = 	 {Proceedings of Machine Learning Research},
  month = 	 {13--19 Jul},
  publisher =    {PMLR},
  url = 	 {https://proceedings.mlr.press/v267/xu25e.html},
}

@INPROCEEDINGS{spikinghybridyolo,
  author={Guo, Mingxin and Xu, Dongjun and Li, Yaoyao and Cheng, Jian and Chen, Liang},
  booktitle={2024 IEEE International Symposium on Circuits and Systems (ISCAS)}, 
  title={Spiking-Hybrid-YOLO for Low-Latency Object Detection}, 
  year={2024},
  volume={},
  number={},
  pages={1-5},
  doi={10.1109/ISCAS58744.2024.10558470}}

@INPROCEEDINGS{addernet,
  author={Chen, Hanting and Wang, Yunhe and Xu, Chunjing and Shi, Boxin and Xu, Chao and Tian, Qi and Xu, Chang},
  booktitle={2020 IEEE/CVF Conference on Computer Vision and Pattern Recognition (CVPR)}, 
  title={{AdderNet: Do We Really Need Multiplications in Deep Learning?}}, 
  year={2020},
  volume={},
  number={},
  pages={1465-1474},
  doi={10.1109/CVPR42600.2020.00154}}

@INPROCEEDINGS{mobilenetv2,
  author={Sandler, Mark and Howard, Andrew and Zhu, Menglong and Zhmoginov, Andrey and Chen, Liang-Chieh},
  booktitle={2018 IEEE/CVF Conference on Computer Vision and Pattern Recognition}, 
  title={{MobileNetV2: Inverted Residuals and Linear Bottlenecks}}, 
  year={2018},
  volume={},
  number={},
  pages={4510-4520},
  doi={10.1109/CVPR.2018.00474}}

@ARTICLE{snnfpga,
  author={Farsa, Edris Zaman and Ahmadi, Arash and Keszocze, Oliver},
  journal={IEEE Transactions on Emerging Topics in Computational Intelligence}, 
  title={{Reconfigurable Digital FPGA Implementations for Neuromorphic Computing: A Survey on Recent Advances and Future Directions}}, 
  year={2025},
  volume={9},
  number={5},
  pages={3210-3232},
  doi={10.1109/TETCI.2025.3551934}}

@INPROCEEDINGS{horowitz,
  author={Horowitz, Mark},
  booktitle={2014 IEEE International Solid-State Circuits Conference Digest of Technical Papers (ISSCC)}, 
  title={{1.1 Computing's energy problem (and what we can do about it)}}, 
  year={2014},
  volume={},
  number={},
  pages={10-14},
  doi={10.1109/ISSCC.2014.6757323}}

@ARTICLE{tcas1,
  author={Lien, Hong-Han and Chang, Tian-Sheuan},
  journal={IEEE Transactions on Circuits and Systems I: Regular Papers}, 
  title={Sparse Compressed Spiking Neural Network Accelerator for Object Detection}, 
  year={2022},
  volume={69},
  number={5},
  pages={2060-2069},
  doi={10.1109/TCSI.2022.3149006}}

@INPROCEEDINGS{iscasdronesnn,
  author={Li, Zehao and Lu, Wenhao and Lu, Yuncheng and Li, Junying and Shi, Yucen and Zheng, Yuanjin and Kim, Tony Tae-Hyoung},
  booktitle={2024 IEEE International Symposium on Circuits and Systems (ISCAS)}, 
  title={An Energy-Efficient Object Detection System in IoT with Dynamic Neuromorphic Vision Sensors}, 
  year={2024},
  volume={},
  number={},
  pages={1-5},
  doi={10.1109/ISCAS58744.2024.10558689}}

@INPROCEEDINGS{aicas,
  author={Silva, D. A. and Shymyrbay, A. and Smagulova, K. and Elsheikh, A. and Fouda, M. E. and Eltawil, A. M.},
  booktitle={2024 IEEE 6th International Conference on AI Circuits and Systems (AICAS)}, 
  title={End-to-End Edge Neuromorphic Object Detection System}, 
  year={2024},
  volume={},
  number={},
  pages={194-198},
  doi={10.1109/AICAS59952.2024.10595906}}

@INPROCEEDINGS{sfod,
  author={Fan, Yimeng and Zhang, Wei and Liu, Changsong and Li, Mingyang and Lu, Wenrui},
  booktitle={2024 IEEE/CVF Conference on Computer Vision and Pattern Recognition (CVPR)}, 
  title={SFOD: Spiking Fusion Object Detector}, 
  year={2024},
  volume={},
  number={},
  pages={17191-17200},
  doi={10.1109/CVPR52733.2024.01627}}

@ARTICLE{spikingvit,
  author={Yu, Lixing and Chen, Hanqi and Wang, Ziming and Zhan, Shaojie and Shao, Jiankun and Liu, Qingjie and Xu, Shu},
  journal={IEEE Transactions on Cognitive and Developmental Systems}, 
  title={SpikingViT: A Multiscale Spiking Vision Transformer Model for Event-Based Object Detection}, 
  year={2025},
  volume={17},
  number={1},
  pages={130-146},
  doi={10.1109/TCDS.2024.3422873}}

@article{ncsnnmnist,
  author    = {Aleksandar Stanojevic and Szymon Wo{\'z}niak and Guillaume Bellec and
               Johannes C. Schemmel and Friedemann Zenke and Wulfram Gerstner},
  title     = {{High-performance deep spiking neural networks with 0.3 spikes per neuron}},
  journal   = {Nature Communications},
  volume     = {15},
  number     = {1},
  pages      = {6793},
  year       = {2024},
  doi        = {10.1038/s41467-024-51110-5},
  url        = {https://doi.org/10.1038/s41467-024-51110-5}
}

@ARTICLE{aodet,
  author={Wang, Xiaoming and Chen, Hao and Chu, Xiangxiang and Wang, Peng},
  journal={IEEE Transactions on Geoscience and Remote Sensing}, 
  title={{AODet: Aerial Object Detection Using Transformers for Foreground Regions}}, 
  year={2024},
  volume={62},
  number={},
  pages={1-11},
  doi={10.1109/TGRS.2024.3407815}}

@ARTICLE{cascadercnn,
  author={Cai, Zhaowei and Vasconcelos, Nuno},
  journal={IEEE Transactions on Pattern Analysis and Machine Intelligence}, 
  title={{Cascade R-CNN: High Quality Object Detection and Instance Segmentation}}, 
  year={2021},
  volume={43},
  number={5},
  pages={1483-1498},
  doi={10.1109/TPAMI.2019.2956516}}

@inproceedings{deformabledetr,
  author    = {Xizhou Zhu and Weijie Su and Lewei Lu and Bin Li and Xiaogang Wang and Jifeng Dai},
  title     = {{Deformable DETR: Deformable Transformers for End-to-End Object Detection}},
  booktitle = {International Conference on Learning Representations (ICLR)},
  year      = {2021},
  url       = {https://openreview.net/forum?id=gZ9hCDWe6ke}
}

@inproceedings{sparsedetr,
  author    = {Byungseok Roh and JaeWoong Shin and Wuhyun Shin and Saehoon Kim},
  title     = {{Sparse DETR: Efficient End-to-End Object Detection with Learnable Sparsity}},
  booktitle = {International Conference on Learning Representations (ICLR)},
  year      = {2022},
  url       = {https://openreview.net/forum?id=RRGVCN8kjim}
}

@ARTICLE{lwuavdet,
  author={Min, Xuanlin and Zhou, Wei and Hu, Rui and Wu, Yinyue and Pang, Yiran and Yi, Jun},
  journal={IEEE Internet of Things Journal}, 
  title={{LWUAVDet: A Lightweight UAV Object Detection Network on Edge Devices}}, 
  year={2024},
  volume={11},
  number={13},
  pages={24013-24023},
  doi={10.1109/JIOT.2024.3388045}}

@ARTICLE{mcayolo,
  author={Chen, Zhihuan and Luo, Aiwen and Ding, Lin and Zheng, Jialu and Huang, Zunkai},
  journal={IEEE Geoscience and Remote Sensing Letters}, 
  title={{A Robust and Efficient Multiscenario Object Detection Network for Edge Devices}}, 
  year={2025},
  volume={22},
  number={},
  pages={1-5},
  doi={10.1109/LGRS.2025.3529830}}

@ARTICLE{tai_dota,
  author={Weng, Weijie and Wei, Mengwan and Ren, Junchi and Shen, Fei},
  journal={IEEE Transactions on Artificial Intelligence}, 
  title={{Enhancing Aerial Object Detection With Selective Frequency Interaction Network}}, 
  year={2024},
  volume={5},
  number={12},
  pages={6109-6120},
  doi={10.1109/TAI.2024.3381096}}

@ARTICLE{tai_small,
  author={Chan, Sixian and Zheng, Jingcheng and Li, Suqiang and Mu, Pan and Bai, Cong},
  journal={IEEE Transactions on Artificial Intelligence}, 
  title={{Adaptive Feature Fusion Based on Double Head Framework for Small Object Detection in Highway Scenarios}}, 
  year={2026},
  volume={7},
  number={8},
  pages={4464-4475},
  doi={10.1109/TAI.2026.3664780}}

@INPROCEEDINGS{orientedrcnn,
  author={Xie, Xingxing and Cheng, Gong and Wang, Jiabao and Yao, Xiwen and Han, Junwei},
  booktitle={2021 IEEE/CVF International Conference on Computer Vision (ICCV)}, 
  title={{Oriented R-CNN for Object Detection}}, 
  year={2021},
  volume={},
  number={},
  pages={3500-3509},
  doi={10.1109/ICCV48922.2021.00350}}

@INPROCEEDINGS{lsknet,
  author={Li, Yuxuan and Hou, Qibin and Zheng, Zhaohui and Cheng, Ming-Ming and Yang, Jian and Li, Xiang},
  booktitle={2023 IEEE/CVF International Conference on Computer Vision (ICCV)}, 
  title={{Large Selective Kernel Network for Remote Sensing Object Detection}}, 
  year={2023},
  volume={},
  number={},
  pages={16748-16759},
  doi={10.1109/ICCV51070.2023.01540}}

@INPROCEEDINGS{pkinet,
  author={Cai, Xinhao and Lai, Qiuxia and Wang, Yuwei and Wang, Wenguan and Sun, Zeren and Yao, Yazhou},
  booktitle={2024 IEEE/CVF Conference on Computer Vision and Pattern Recognition (CVPR)}, 
  title={{Poly Kernel Inception Network for Remote Sensing Detection}}, 
  year={2024},
  volume={},
  number={},
  pages={27706-27716},
  doi={10.1109/CVPR52733.2024.02617}}

@ARTICLE{msresnet,
  author={Hu, Yifan and Deng, Lei and Wu, Yujie and Yao, Man and Li, Guoqi},
  journal={IEEE Transactions on Neural Networks and Learning Systems}, 
  title={{Advancing Spiking Neural Networks Toward Deep Residual Learning}}, 
  year={2025},
  volume={36},
  number={2},
  pages={2353-2367},
  doi={10.1109/TNNLS.2024.3355393}}

@article{dgrdet,
title = {{DGRDet: Dynamic Gaussian Receptive Field Encoding-Based Spiking Neural Networks for Remote Sensing Object Detection}},
journal = {Computers, Materials and Continua},
volume = {88},
number = {2},
year = {2026},
issn = {1546-2218},
doi = {https://doi.org/10.32604/cmc.2026.078314},
url = {https://www.sciencedirect.com/science/article/pii/S1546221826005953},
author = {Li Chen and Fan Zhang and Guangwei Xie and Yanzhao Gao and Xiaofeng Qi and Mingqian Sun}
}

@Article{makbrnet,
AUTHOR = {Niu, Ge and Yang, Xiaolong and Wang, Xinhui and Liu, Yong and Cao, Lu and Yin, Erwei and Guo, Pengyu},
TITLE = {{MAK-BRNet: Multi-Scale Adaptive Kernel and Boundary Refinement Network for Remote Sensing Object Detection}},
JOURNAL = {Applied Sciences},
VOLUME = {16},
YEAR = {2026},
NUMBER = {1},
ARTICLE-NUMBER = {522},
URL = {https://www.mdpi.com/2076-3417/16/1/522},
ISSN = {2076-3417},
DOI = {10.3390/app16010522}
}

@ARTICLE{arsdetr,
  author={Zeng, Ying and Chen, Yushi and Yang, Xue and Li, Qingyun and Yan, Junchi},
  journal={IEEE Transactions on Geoscience and Remote Sensing}, 
  title={{ARS-DETR: Aspect Ratio-Sensitive Detection Transformer for Aerial Oriented Object Detection}}, 
  year={2024},
  volume={62},
  number={},
  pages={1-15},
  doi={10.1109/TGRS.2024.3364713}}

@ARTICLE{ao2detr,
  author={Dai, Linhui and Liu, Hong and Tang, Hao and Wu, Zhiwei and Song, Pinhao},
  journal={IEEE Transactions on Circuits and Systems for Video Technology}, 
  title={{AO2-DETR: Arbitrary-Oriented Object Detection Transformer}}, 
  year={2023},
  volume={33},
  number={5},
  pages={2342-2356},
  doi={10.1109/TCSVT.2022.3222906}}

@INPROCEEDINGS{fcos,
  author={Tian, Zhi and Shen, Chunhua and Chen, Hao and He, Tong},
  booktitle={2019 IEEE/CVF International Conference on Computer Vision (ICCV)}, 
  title={{FCOS: Fully Convolutional One-Stage Object Detection}}, 
  year={2019},
  volume={},
  number={},
  pages={9626-9635},
  doi={10.1109/ICCV.2019.00972}}

@ARTICLE{s2anet,
  author={Han, Jiaming and Ding, Jian and Li, Jie and Xia, Gui-Song},
  journal={IEEE Transactions on Geoscience and Remote Sensing}, 
  title={{Align Deep Features for Oriented Object Detection}}, 
  year={2022},
  volume={60},
  number={},
  pages={1-11},
  doi={10.1109/TGRS.2021.3062048}}

@INPROCEEDINGS{roitrans,
  author={Ding, Jian and Xue, Nan and Long, Yang and Xia, Gui-Song and Lu, Qikai},
  booktitle={2019 IEEE/CVF Conference on Computer Vision and Pattern Recognition (CVPR)}, 
  title={{Learning RoI Transformer for Oriented Object Detection in Aerial Images}}, 
  year={2019},
  volume={},
  number={},
  pages={2844-2853},
  doi={10.1109/CVPR.2019.00296}}

@INPROCEEDINGS{oreppoints,
  author={Li, Wentong and Chen, Yijie and Hu, Kaixuan and Zhu, Jianke},
  booktitle={2022 IEEE/CVF Conference on Computer Vision and Pattern Recognition (CVPR)}, 
  title={{Oriented RepPoints for Aerial Object Detection}}, 
  year={2022},
  volume={},
  number={},
  pages={1819-1828},
  doi={10.1109/CVPR52688.2022.00187}}

@inproceedings{r3detkld,
  title={{Learning High-Precision Bounding Box for Rotated Object Detection via Kullback-Leibler Divergence}},
  author={Yang, Xue and Yang, Xiaojiang and Yang, Jirui and Ming, Qi and Wang, Wentao and Tian, Qi and Yan, Junchi},
  booktitle={Advances in Neural Information Processing Systems},
  volume={34},
  pages={18381--18394},
  year={2021}
}

@ARTICLE{s3det,
  author={Chen, Li and Zhang, Fan and Xie, Guangwei and Gao, Yanzhao and Qi, Xiaofeng and Sun, Mingqian},
  journal={Frontiers of Information Technology \& Electronic Engineering}, 
  title={{S3Det: A Fast Object Detector for Remote Sensing Images Based on Artificial to Spiking Neural Network Conversion}}, 
  year={2025},
  volume={26},
  number={5},
  pages={713-727},
  doi={10.1631/FITEE.2400594}}


 





\end{document}